\documentclass[10pt,twocolumn,letterpaper]{article}

\usepackage{iccv}
\usepackage{times}
\usepackage{epsfig}
\usepackage{graphicx}
\usepackage{amsmath}
\usepackage{amssymb}

\usepackage{multirow}
\usepackage{booktabs}

\usepackage{relsize}
\usepackage{color}
\usepackage[usenames,dvipsnames]{xcolor}
\usepackage{caption}
\usepackage{subcaption}
\usepackage[accsupp]{axessibility} 

\usepackage{pifont}

\DeclareMathOperator*{\argmax}{arg\,max}

\usepackage[pagebackref=true,breaklinks=true,colorlinks,bookmarks=false]{hyperref}

\iccvfinalcopy 

\def\iccvPaperID{7923} 

\ificcvfinal\fi

\begin{document}

\title{The Center of Attention: Center-Keypoint Grouping via  Attention for Multi-Person Pose Estimation}

\author{Guillem Brasó \qquad  \qquad Nikita Kister \qquad  \qquad  Laura Leal-Taixé\\
Technical University of Munich\\
{\tt\small \{guillem.braso,\qquad   n.kister, \qquad  lealtaixe\}@tum.de}
}

\maketitle
\ificcvfinal\thispagestyle{empty}\fi

\begin{abstract}
We introduce CenterGroup, an attention-based framework to estimate human poses from a set of identity-agnostic keypoints and person center predictions in an image. 
Our approach uses a transformer to obtain context-aware embeddings for all detected keypoints and centers and then applies multi-head attention to directly group joints into their corresponding person centers. While most bottom-up methods rely on non-learnable clustering at inference, CenterGroup uses a fully differentiable attention mechanism that we train end-to-end together with our keypoint detector. As a result, our method obtains state-of-the-art performance with up to 2.5x faster inference time than \makebox[\textwidth][s]{competing bottom-up approaches. Our code is available at\\} \url{https://github.com/dvl-tum/center-group} 
\end{abstract}
\vspace{-15pt}

\section{Introduction}

Localizing the anatomical 2D keypoints of all humans in an image is a fundamental task in computer vision, with the ability to enable progress in applications such as virtual reality, human computer interaction, and human behavior analysis. It is also a common key component of algorithms for tasks such as action recognition \cite{skeleton_gcns, shiftGCN}, multi-object tracking \cite{mot_joints, jcc}, and generative models \cite{dance_pose, person_gan}. 


Current methods typically follow one of two paradigms: \textit{bottom-up} and \textit{top-down}. Top-down approaches \cite{cascpyramid, rmpe, mask_rcnn, coarsefine, crowdpose, inthewild, simple_baselines, hrnet_original, integral} divide the problem into two subtasks: (i) bounding box detection for all persons in the image, and (ii) joint localization for each person individually. Despite their success in some benchmarks \cite{mpii_pose, posetrack, coco_dataset}, these two-step approaches lack efficiency due to their need to use a separate object detector, and their performance tends to degrade severely under heavy occlusions \cite{crowdpose}.
Bottom-up methods \cite{openpose, deepcut, deepercut,hourglass_ae, personlab, hrnet, hgg} follow a different approach, as they first detect identity-agnostic keypoints, and then group them into separate poses. Their lack of reliance on external object detectors and their ability to operate jointly over the entire set of keypoints in the image has allowed them to outperform top-down approaches in benchmarks where occlusions are common \cite{crowdpose}. While recent work has significantly advanced the ability of bottom-up methods to accurately predict identity-free keypoints \cite{hrnet}, current grouping algorithms still face significant drawbacks: since they generally rely on optimization algorithms, they cannot be trained end-to-end, and are often slow.
\begin{figure}[t]
\begin{center}
   \includegraphics[width=1.\linewidth]{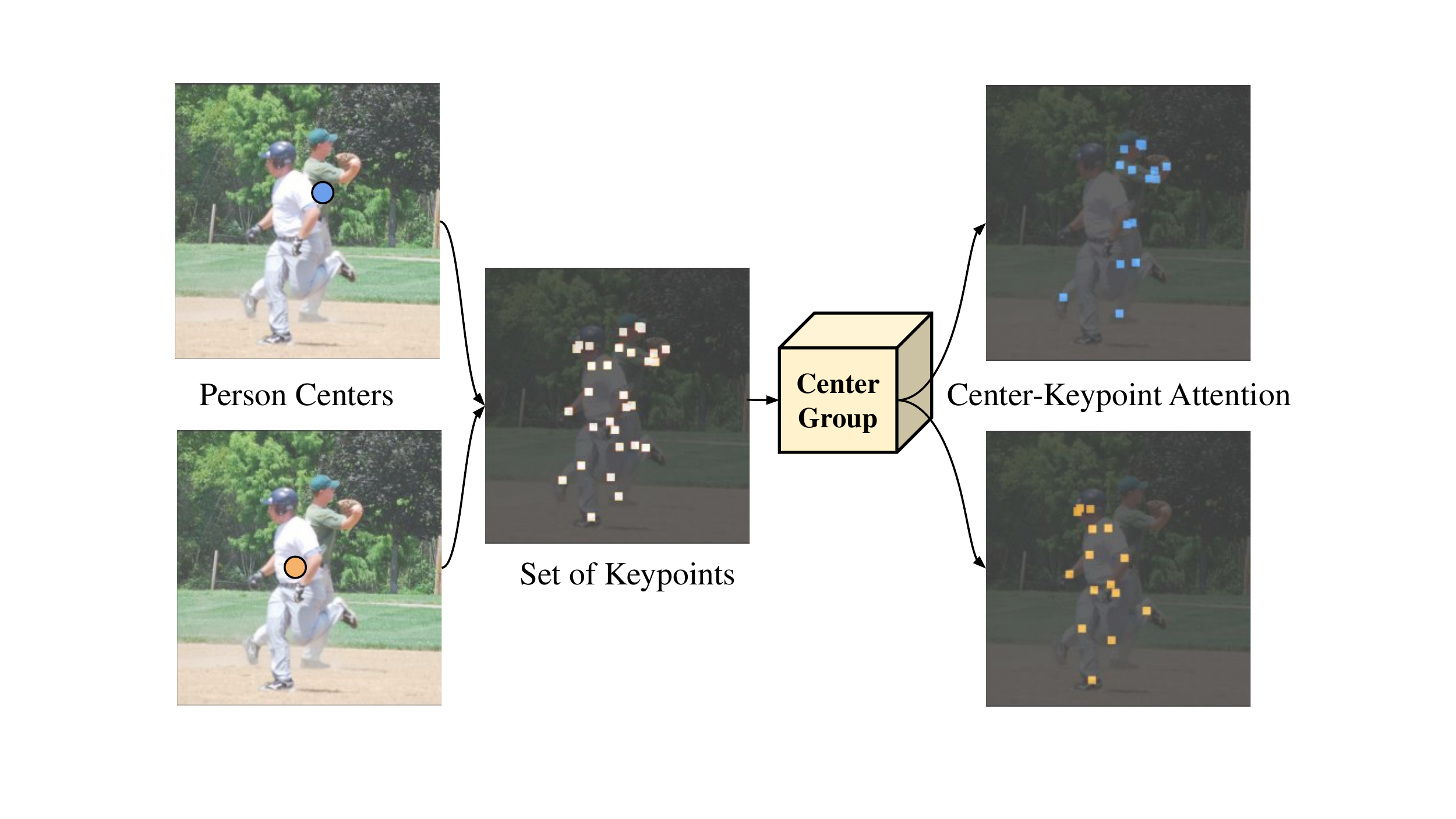}
\end{center}
\vspace{-10pt}
   \caption{
      Given a set of predicted identity-agnostic keypoints and person centers, CenterGroup learns to assign keypoints into their corresponding centers with attention.}
   
   
\label{fig:teaser}
\label{fig:onecol}
\vspace{-10pt}
\end{figure}

The keypoint grouping task can be formulated as a graph optimization problem in which nodes represent keypoints, and edge weights, which can be learned, represent their likelihood of belonging to the same human pose.  Approaches ranging from integer linear programming \cite{deepcut, deepercut, arttrack}, heuristics \cite{hourglass_ae, personlab, openpose} or graph clustering \cite{hgg} are then used to find the correct assignment. 
%
%
A common problem of bottom-up methods is that their learning objectives are poorly aligned with the real inference procedure: they learn affinities between keypoints but, at test time, grouping is performed by a separate algorithm which is not differentiable \textit{per se}.



One-shot methods are an efficient alternative \cite{spm, centernet, pointsetanchors} to optimization-based bottom-up methods.
Their general formulation consists in regressing a \textit{root node} location per person, and then predicting offsets to  keypoint  locations. Since they are able to avoid the optimization-based grouping stage, they are significantly faster than their counterparts. However, given the inherent difficulty of predicting offsets under occlusions and scale variation, they are also significantly less accurate, and therefore have to rely on additional postprocessing techniques to obtain competitive performance \cite{spm, centernet, pointsetanchors}. 


We propose to tackle the limitations of current bottom-up grouping and one-shot algorithms with a novel framework based on attention. Instead of regressing offsets from a set of center nodes, our proposed CenterGroup uses attention to search for the best match between person centers and keypoints over the entire image. Our method retains the ability of bottom-up approaches to precisely predict keypoints from heatmaps, while maintaining the efficiency of one-shot methods. Furthermore, unlike standard bottom-up methods, CenterGroup does not require any test-time optimization and is end-to-end trainable. 
%
%

More specifically, we first obtain proposals for person centers and identity-agnostic keypoints via heatmap regression. 
We then feed centers and keypoints to a transformer \cite{transformer} to encode contextual information into their updated embeddings.
Finally, the embeddings are used in a simple keypoint grouping scheme which maximizes the attention scores between person centers and keypoints belonging to the same pose. At test time, we extract poses by assigning to centers those keypoints with the corresponding highest attention score. Due to the simplicity of our grouping algorithm and the parallel nature of attention computation, CenterGroup is 2.5x faster than the current state-of-the-art bottom-up method~\cite{hrnet}, while having better performance. 

%
%
%
%
Overall, we make the following \textbf{contributions}:
\begin{itemize}
\item 
We propose to tackle the pose estimation problem by grouping keypoints and person center predictions with a multi-head attention formulation that allows to train the model in an end-to-end fashion. 
\item We use a transformer to encode dependencies between bottom-up detected keypoints and centers to obtain context-enhanced embeddings, efficiently boosting the performance of our proposed grouping scheme.
\item We achieve state-of-the-art results within an end-to-end framework that yields a speedup increase of up to 2.5x with respect to state-of-the-art \cite{hrnet}.
\end{itemize}


\section{Related Work}
\textbf{Top-down methods.} Top-down methods \cite{cascpyramid, rmpe, mask_rcnn, coarsefine, crowdpose, inthewild, simple_baselines, hrnet_original, integral, gcpnn, dark, devil, delicate, hourglass_orig, posefix, channel-spatial, peek, prob-graph, top-down-attn} split the task into two steps.
They first apply a person detector on the image and then perform single person pose estimator for each detected image region which is given by the bounding boxes.
While being particularly strong at handling scale variaion, these methods struggle in cases of occlusion. To address these limitations, previous work has explored refining poses by exploiting the graph structure of the human skeletons with additional modules such as graph networks \cite{gcpnn, bulat2016human, peek} or probabilistic graphical models\cite{prob-graph}. While being more robust, these methods still rely on external detectors, and therefore cannot recover from missing  boxes. 

\textbf{Bottom-up methods}. Bottom-up methods \cite{openpose, deepcut, deepercut,hourglass_ae, personlab, hrnet, hgg, pifpaf} start by detecting identity-free keypoints over the entire image. 
In a second step, a grouping algorithm assembles poses using pairwise similarity scores between keypoints.
To predict these similarity scores, DeepCut and Person-Lab \cite{deepcut, deepercut, personlab} predict offset fields that link joints belonging to the same person.
Openpose and PifPaf \cite{openpose, pifpaf} predict part affinity fields which resemble human limbs and encode the position and orientation between pairs of keypoints.
Associative embeddings \cite{hourglass_ae}  are a popular approach  currently used by state-of-the-art~\cite{hrnet}. They predict an embedding for every detected keypoint from convolutional features, and then use their pairwise euclidean distances as similarity scores. For all these methods, grouping is done by either graph partitioning \cite{deepcut, deepercut, posetrack, arttrack, hgg, local-joint} or heuristic greedy parsing \cite{hourglass_ae, openpose, personlab}. 
HGG\cite{hgg} makes progress towards learning to group keypoints by using a graph neural network~\cite{graph_net_original} on top of associative embeddings and training an edge and node classifier to hierarchically predict which keypoints belong together. While its graph network is trainable, it still relies on an external non-differentiable clustering algorithm~\cite{graclus} for grouping.
CenterGroup does not need this clustering step and, instead, it uses attention as a form of learnable keypoint grouping.

\textbf{One-shot methods}. One-shot methods\cite{spm, centernet, pointsetanchors} avoid the grouping task by directly regressing keypoint locations from a set of predicted centers \cite{spm, centernet} or anchors\cite{pointsetanchors}. Both SPM\cite{spm} and CenterNet\cite{centernet} regress offsets at center locations to regress each person's joints.
In addition, \cite{centernet} predicts keypoint heatmaps as standard bottom-up methods. It then combines both predictions  by heuristically matching offsets to their closest predicted joints. While being more efficient than grouping-based approaches, this heuristic still does not perform on par with them, and suffers from the same problem of not being end-to-end learnable.
\textbf{Transformers and attention}. Transformers were initially introduced for machine translation, and became recently popular for computer vision tasks ranging from image classification \cite{vit, igpt}, object detection \cite{detr, ddetr}, semantic segmentation \cite{deeplab-former, vistr}, video processing \cite{sttn, masked-former}, image generation \cite{image-transformer}, and hand pose estimation \cite{hand-former, hot-net}. They employ self-attention layers to model relations between entities in a global context. 
Their use for human pose estimation is still relatively unexplored: \cite{15keypoints} employs transformer for human pose tracking,  \cite{ lift-former, epi-former} apply them to estimate 3D human poses and \cite{transpose} use a transformer based architecture for explainable single person pose estimation. 

\vspace{-0.2cm}
\section{Background: Multi-Head Attention} \label{section:background}
Our model uses multi-head attention and a transformer as main tools to perform grouping. Therefore, we start by providing a brief review of these techniques.

Multi-Head Attention (MHA) \cite{transformer}, the core component of the transformer model, aims at obtaining contextual representations from an unordered set of vectors by letting each vector \textit{attend} over multiple representation subspaces of a (possibly different) set of vectors. 
More precisely, given a set of $n$ $d-$dimensional \textit{query} feature vectors, $Q:=\{q_i \in \mathbb{R}^{ d}\}_{i=1}^n$, and a set of $m$ pairs of \textit{key} and \textit{value} vectors, $K:=\{k_j \in \mathbb{R}^{d}\}_{j=1}^m$ and $V:=\{v_j \in \mathbb{R}^{d}\}_{j=1}^m$\footnote{For notational simplicity, we assume queries, keys and values have the same dimensinality.}, MHA updates the query embeddings by linearly projecting the concatenation of $h$ attention heads:
\begin{equation}
\text{MHA}(Q, K, V)= \text{concat}(\text{head}_1, \dots, \text{head}_h)W_O,
\end{equation}

where $W_O\in \mathbb{R}^{(d_H*h)\times d}$ is a learnable matrix, and $d_H$ is the dimensionality of each attention head. Each attention head computes, at every index $l\in \{1, \dots n\}$:
\begin{equation}
(head_i)_l= \sum_{j=1}^m \text{attn}_i(q_l, k_j) W_i^Vv_j, 
\end{equation}
where the attention scores are computed as softmax-normalized\footnote{Transformers use \textit{scaled} dot product attention, which means that they normalize softmax outputs with the dimensionality of the projected embeddings, $d_H$. However, we omit this term as we will not use it in the remaining of the paper.} dot-products between keys and queries:
\begin{equation}
\text{attn}_i(q_l, k_j):=\frac{\exp((W_i^Qq_l)^T(W_i^Kk_j))}{\sum_{t=1}^m\exp((W_i^Qq_l)^T(W_i^Kk_t))}
\end{equation}

where, $W_i^Q,W_i^K, W_i^V\in\mathbb{R}^{d\times d_H}$ are learnable projection matrices.


Whenever these sets of key, query and values are the same, i.e., $Q = K = V$, one refers to \textit{self-attention}, which is the core component of the transformer encoder architecture. Overall, transformer encoders are formed by stacking blocks of an initial layer of self-attention with skip-connection and layer normalization \cite{layernorm}, followed by a feed-forward network and a second instance of layer normalization. For completeness, we provide a more detailed explanation of their architecture in the supplementary material.


\section{Problem Formulation}
\begin{figure*}[ht]
\begin{center}
\vspace{-0.5cm}
\includegraphics[width=0.945\linewidth]{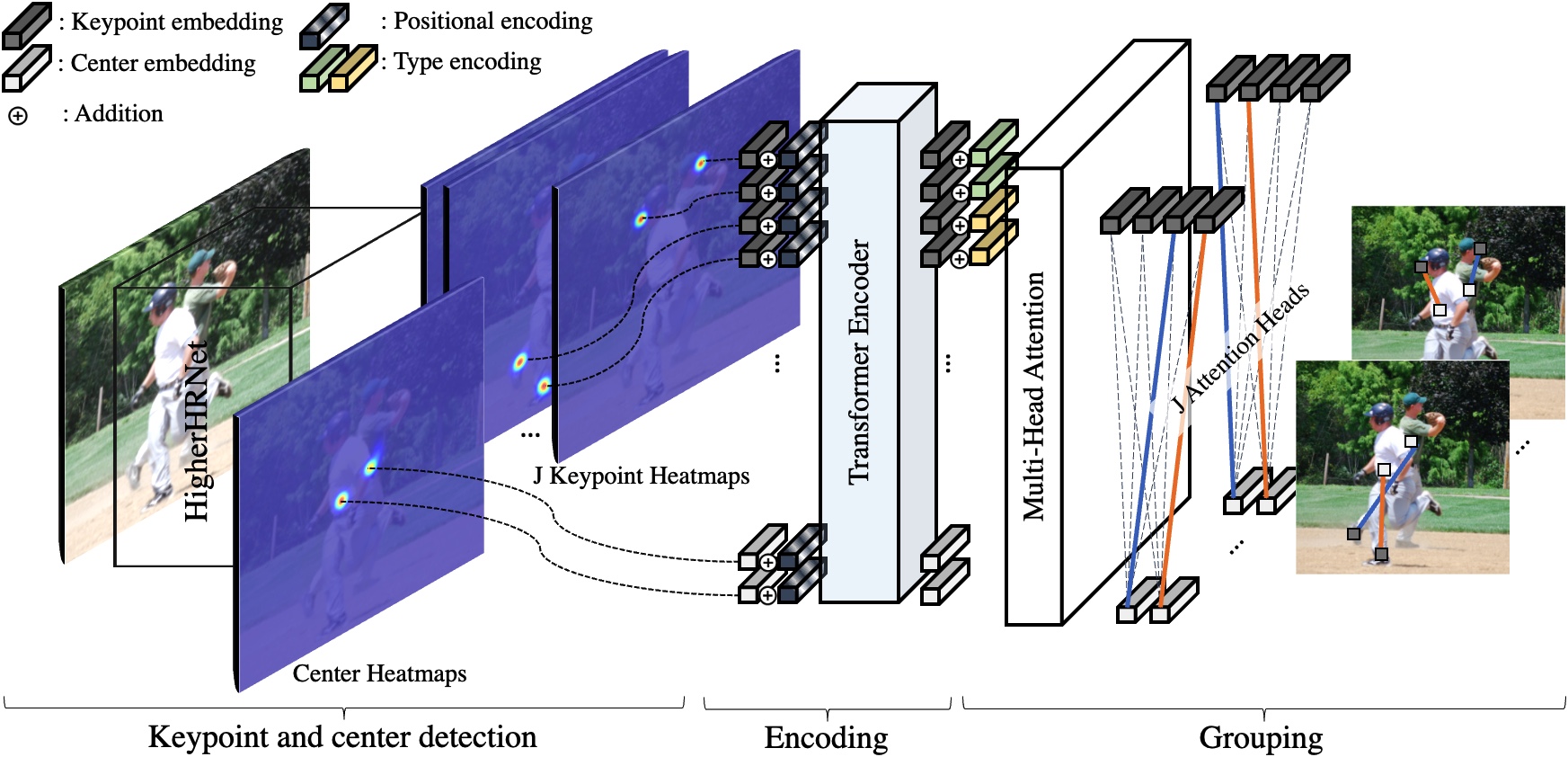}
%
%
\captionof{figure}{Our method receives a single RGB image as input and predicts a set of identity-agnostic keypoints and person centers with a HigherHRNet\cite{hrnet} by heatmap regression. It then extracts  features from our CNN's last layer, augments them with positional encodings and feeds them to a transformer encoder that returns context-aware embeddings for every keypoint and center. These embeddings are then fed to an attention module that predicts which joints correspond to each center.}
\label{fig:pipeline}
\vspace{-22pt}
\end{center}
\end{figure*}

We first  provide an overview of the general formulation of our method and introduce notation. 

\subsection{Problem Statement}
Given an input image,  we aim to obtain the set of poses $\mathcal{P}$ corresponding to all persons in the image. Let $J$ be the number of joints being considered. Each pose can be uniquely determined by the 2D location and visibility of its $J$ joints. Formally, for every pose $p\in \mathcal{P}$, we refer to its joint locations as $\text{loc}_p^i\in \mathbb{R}^2$ for every $i\in \{1, \dots, J\}$. We denote the visibility of each joint of pose $p$ as $\text{vis}_p^i\in \{0, 1\}$, and assign it 1 whenever the joint is visible, and 0 otherwise. 

Our approach operates over a set of predicted identity-agnostic joint keypoints in the image, which we refer to as $\mathcal{K}$. Each keypoint $k\in \mathcal{K}$ can be identified by its 2D location $\text{loc}_k\in \mathbb{R}$ and its predicted type $\text{type}_k\in \{1, \dots, J\}$. Our method also predicts an additional set of targets corresponding to the center locations of persons in the image. We denote as $\mathcal{C}$ the set of detected person centers.

\subsection{Grouping Keypoints and Centers}
Standard bottom-up methods learn a similarity score $\text{sim}(k_1, k_2)$ for every pair of detected keypoints $k_1, k_2\in \mathcal{K}$, and use them to form poses by clustering keypoints that are most similar. One-shot methods, instead, directly use predicted person centers and regress displacement offsets from centers to joint locations to avoid expensive grouping. 

Inspired by these approaches, we propose to perform human pose estimation by learning  similarity score $\text{sim}_i(c, k)\in \mathbb{R}^+$ between every pair of detected person center $c\in \mathcal{C}$ and keypoints $k\in \mathcal{K}$ and type $i\in \{1, \dots, J\}$.  By being able to estimate the similarity between center nodes and keypoints, as opposed to the similarity between pairs of keypoints, we are able to reduce the complexity of the grouping task significantly. Instead of requiring a graph clustering algorithm, we formulate the grouping task as a simple nearest-neighbor search problem. Namely, for every predicted center $c\in \mathcal{C}$, we obtain its corresponding pose by retrieving the locations of its most similar detected keypoint $k^*\in \mathcal{K}$ of a target type $i\in \{1, \dots, J\}$ according to $\text{sim}_i$. Formally, the predicted location  $\widehat{\text{loc}}_c^i$ of joint type $i$ for center $c$ can be obtained as $\widehat{\text{loc}}_c^i = \text{loc}_{k^*}$, where
\begin{equation} \label{eq:closest_of_type}
    k^* = \argmax_{k\in \mathcal{K}} \text{sim}_i(c, k)
\end{equation}
Since our approach operates directly over the set of detected keypoint locations, there is no need for additional postprocessing to obtain precise joint locations, unlike offset-based  methods \cite{centernet, spm}.

\subsection{Attention as Differentiable Keypoint Selection} \label{attention_selection}
The main drawback from the aforementioned procedure is that it is not end-to-end trainable, as it involves an $\argmax$ operation over the detected keypoints. We circumvent this issue by formulating the nearest neighbor search task as a differentiable attention mechanism. To do so, we treat person centers as our set of queries, and keypoints as our set of keys, and obtain their similarity scores by computing their dot-product in a learned embedding space for every joint type $i\in \{1, \dots, J\}$. We then normalize the scores with a softmax operator to replace the non-differentiable $\argmax$. The resulting coefficients are used during training to directly predict keypoint locations for every joint type $i\in \{1, \dots, J\}$ and every person center $c$ as:
\begin{equation} \label{problem_form:train_pred}
\widehat{\text{loc}}_c^i := \sum_{k\in \mathcal{K}}\text{attn}_i(c, k)\text{loc}_k
\end{equation}
where  $\text{loc}_k$ are the coordinates of the detected keypoint $k$. 

Predictions resulting from equation \ref{problem_form:train_pred} can then be minimized by directly computing their $L1$ loss with respect to the ground truth location.
%
Note that since the detected keypoint coordinates are fixed in Equation \ref{problem_form:train_pred}, in order to minimize the loss our network needs to assign the highest attention score to the keypoints which location is closest to the ground truth coordinates $\text{loc}_c^i$. Moreover, in the limit, when $\max_{k\in \mathcal{K}}\text{attn}_i(c, k)=1$, Equation \ref{problem_form:train_pred} becomes equivalent to just computing a standard $\argmax$. Therefore, the attention coefficients act as a differentiable mechanism for selecting keypoints from person centers based on their dot-product similarity. 
This procedure still allows us to use the simple $\argmax$ operator at test-time to efficiently retrieve keypoints from centers as in Equation 
\ref{eq:closest_of_type}.

\section{Method }
We exploit the formulation described in the previous section within an end-to-end bottom-up pipeline for pose estimation. In this section, we first provide a general overview of it, and then explain each of its components in detail. 

\subsection{Overview}
Our method, CenterGroup, consists of three main stages, which are summarized in Figure \ref{fig:pipeline}:
\begin{enumerate}
    \item \textbf{Keypoint and center detection.} The location of identity-agnostic keypoints and person centers is obtained by heatmap regression following HigherHRNet \cite{hrnet}. 
    The output is a variable number of high-scoring joint and person center detections.
    \item \textbf{Encoding keypoints and centers.} For every detected keypoint and center, we extract features from a CNN backbone, and augment them with additional embeddings encoding their spatial position. These embeddings are fed to a transformer~\cite{transformer}, yielding updated embeddings with enhanced contextual information.
    \item \textbf{Keypoint grouping.} We use the embeddings obtained from the previous stage and compute  dot-product attention scores between person centers and keypoints, and normalize them in order to obtain a soft-assignment between persons and keypoints. Additionally, we use the transformer embeddings to classify center nodes into true and false positives, and determine the visibility of each keypoint. 
\end{enumerate}
\subsection{Keypoint and Center Detection} \label{section:detection}
In the first stage of our pipeline, we start by detecting identity agnostic-keypoints and person centers.


\noindent{\bf{Heatmap regression}}. We follow HigherHRNet \cite{hrnet} to obtain identity-agnostic keypoint proposals for each of $J$ joint types being considered. 
HigherHRNet uses an HRNet \cite{hrnet_original}  backbone, followed by two keypoint prediction heads that regress heatmaps at 1/4 and 1/2 of the original image scale for every joint type. Heatmaps are trained to follow a gaussian distribution centered at ground truth keypoint locations. 
During training, both heatmaps are supervised independently with a minimum-squared error loss. At inference, heatmaps are upsampled and aggregated to obtain a single heatmap at full  image  resolution. 

\noindent{\bf{Person centers}}. In addition to joints, we regress a new heatmap corresponding to person centers, also at resolutions 1/4 and 1/2. 
 Following \cite{spm}, given a ground truth pose $p\in \mathcal{P}$ with joint locations $\{\text{loc}_p^i\in \mathbb{R}^2\}_{i = 1}^{J}$, the location of its center is computed as the center of mass of the visible joints, i.e., 
\begin{equation}
\text{loc}_p:=\frac{1}{
N_p}\sum_{\text{vis}_p^i=1}  \text{loc}_p^i. 
\end{equation}
where $N_p:=\sum_{i=1}^J \text{vis}_p^i$ is the number of visible joints in pose $p$. Note that we identify the pose location as that of its center, and hence write $\text{loc}_p$.

\subsection{Encoding Keypoints and Centers}
Given the set of predicted keypoints and centers from the first stage, our goal is to obtain discriminative embeddings encoding contextual information. These embeddings will then be used in our grouping module in order to predict associations among keypoints and centers. Hence, it is desirable for them to encode global context.
Towards this end, we use a transformer encoder that yields updated embeddings for every keypoint and center.
 
\noindent{\bf{Initial features.}} We add one additional residual block \cite{resnet} to our backbone's last feature map at 1/4 of the original resolution. For every detected keypoint and center, we obtain an initial embedding vector by extracting the vector at its corresponding location from the resulting feature map, and feed it to a two-layer Multi-Layer Perceptron (MLP) to project it onto a higher dimensionality. 

\noindent{\bf{Positional encodings.}} CNN features struggle to encode the position of different keypoints \cite{coordconv}. However, spatial information offers an important cue for keypoint grouping. Hence, similarly to previous work \cite{detr, ddetr, vit, vistr}, we use fixed sinusoidal features encoding the absolute $x$ and $y$ axis locations at different frequencies. As a result, we obtain a new vector of dimensionality $d$ and sum it element-wise to the initial features of every detected keypoint and center. 
  
 \noindent{\bf{Transformer encoder.}} In order to encode global context among every detected person and keypoint, we take their initial features, augmented with positional encodings, and feed them to a transformer encoder. As a result, we obtain updated embeddings $h_k$ and $h_c$ for every detected keypoint $k\in \mathcal{K}$ and center $c\in \mathcal{C}$. Our transformer architecture follows the one described in Section \ref{section:background}, and is described in detail in the supplementary material.
\begin{figure}[t]
\begin{center}
   \includegraphics[width=1.\linewidth]{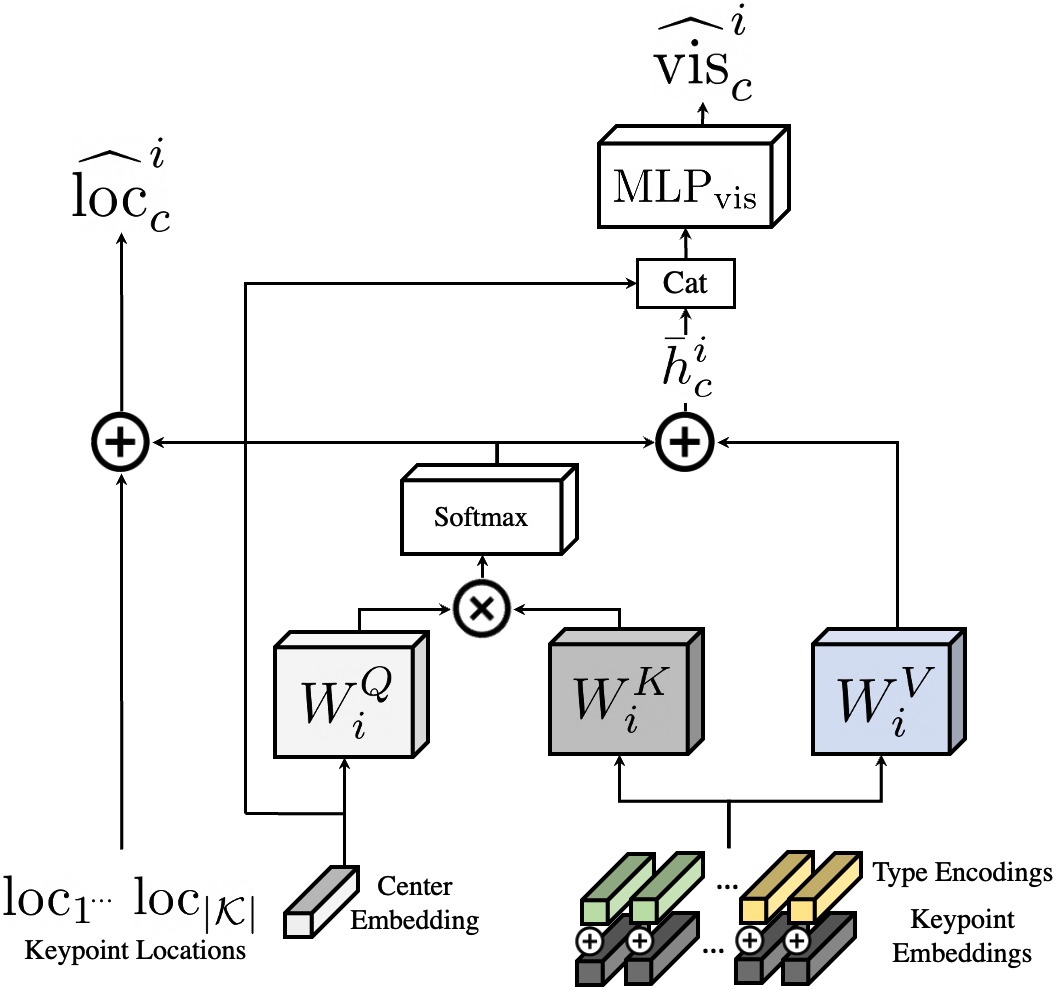}
\end{center}
   \caption{Overview of our Grouping Module. Attention between keypoint and center embeddings is used in order to predict joint locations and visibility for a given center. }
\label{fig:routing}
\vspace{-10pt}
\end{figure}

\subsection{Keypoint Grouping}\label{section:routing}
In the last stage of our pipeline, we construct poses by using the embeddings produced by our transformer to determine which keypoints belong to which person centers via pairwise attention scores. As explained in Section \ref{attention_selection}, we use attention as a differentiable approximation of keypoint selection from centers. In addition, we predict two additional targets for every center: the visibility of each of their keypoints, and the probability that they represent a true pose. This module is summarized in Figure \ref{fig:routing}. 

\noindent{\bf{Classifying centers}}. We start by identifying which predicted centers locations correspond to the ground truth poses by matching them based on their locations\footnote{More details are provided in the supplementary material.}. 
As a result, each predicted center is labelled with a binary target $y^c_{\text{center}}$, set to 1 if the center is matched and 0 otherwise. 
We then use a small multi-layer perceptron to classify the center embeddings produced by our transformer, $h_c$, and supervise the resulting prediction $\text{MLP}_{\text{center}}(h_c)$ with a focal loss~\cite{focal_loss}. 

\noindent{\bf{Predicting joint locations}}. For every predicted center $ c\in \mathcal{C}$ such that $y^c_{\text{center}}=1$, we aim to predict the 2D coordinates of each of its joints $\widehat{\text{loc}}_c^i$ of every type $i\in \{1, \dots, J\}$. 
For each joint type $i\in\{1, \dots, J\}$, we define a pair of learnable projection matrices $W_i^K$ and $W_i^Q$, and a learned \textit{type encoding} vector $\phi_i\in \mathbb{R}^d$. The goal of the projection matrices is to map the center embeddings $h_c$ and keypoint embeddings $h_k$, into a discriminative representation in which their dot-product will encode their likelihood being a good match for type $i$.  We compute their similarity as: 
\begin{equation} \label{eq:sim}
    \text{sim}_i(c, k) = (W_i^Qh_c)^T W_i^K(h_k + \phi_{\text{type}_k})
\end{equation}
Note that the learned embedding $\phi_{\text{type}_k}$ is added to the keypoint embedding $h_k$ before the multiplication. Its goal is to encode the initial type predicted by our keypoint detector for keypoint $k$. Intuitively, when searching for joints of a target type $i$, it is desirable for our network to still be able to consider joints of all predicted types in order to recover from type errors made by the keypoint detector. For instance, for a target type $i$, such as left ankle, some predicted types by the detector (e.g., right ankle) are more likely to be better matching candidates than others (e.g. nose). By using the learnable encoding $\phi_{\text{type}_k}$ for each keypoint before computing the projected embedding with $W^K_i$, we allow our network to explicitely account for the relationships between the target type $i$, and the predicted type of $k$, $\text{type}_k$, in a learnable manner. 

With the similarity scores from Equation \ref{eq:sim}, the final attention scores are computed by normalizing them with a softmax operation over the entire set of keypoints:
\begin{equation}
    \text{attn}_i(c, k) =\frac{\exp{(\text{sim}_i(c, k))}}{\sum_{\bar{k}\in \mathcal{K}} \exp ({\text{sim}_i(c, \bar{k})})}
\end{equation}

Finally, we obtain the predicted locations as in Eq. \ref{problem_form:train_pred}:
\begin{equation} \label{eq:loc_pred}
\widehat{\text{loc}}^i_c = \sum_{k\in \mathcal{K}} \text{attn}_i(c, k)\text{loc}_k,
\end{equation}
and supervise them with an $L$1 loss as:
\begin{equation} 
\mathcal{L}_{\text{loc}}= \sum_{c\in \mathcal{C} |y^c_{\text{center}}=1} \sum_{vis_c^i=1}  |\widehat{\text{loc}}^i_c- \text{loc}^i_c|,
\end{equation}
where the learnable ground locations of every joint for center $\text{loc}^i_c$ are those of its matched ground truth pose.

Overall, this procedure can be interpreted as an instance of an attention head in which center and keypoint embeddings are \textit{queries} and \textit{keys}, respectively, and keypoint locations act as \textit{values}. Note that we use different matrices $W^K_i$ and $W^Q_i$ for every target type $i\in \{1, \dots, J\}$, which is equivalent to having $J$ different heads.

\noindent{\bf{Predicting keypoint visibility}}. One drawback of the attention mechanism we have described is that, due to the softmax normalization, it may still predict high attention scores between centers and keypoints for a target type $i$ even if a given center has no corresponding visible keypoint of that type in the image. We address this problem by exploiting the attention mechanism to explicitly classify whether the predicted keypoints are visible. To do so, we introduce an additional projection matrix for every head $W_i^V$, and reuse the type encodings and attention scores already computed to predict locations to compute a weighted aggregation analogous to that in Equation \ref{eq:loc_pred}: 
\begin{equation}
    \bar{h}_c^i = \sum_{k\in \mathcal{K}} \text{attn}(c, k)W_i^V(h_k + \phi_{\text{type}_k})
\end{equation}
we then concatenate $\bar{h}_c^i$ and $h_c$, and classify the resulting vector with an additional multilayer perceptron, $\text{MLP}_{\text{vis}}$ as either visible or not visible. We supervise the result with a focal loss\footnote{This loss is only computed whenever the given joint in the predicted center is labelled as not visible in the ground truth, or the predicted keypoint has is has small euclidean distance with respect to the with the ground truth keypoint.}. 
Intuitively, whenever keypoints are not visible, the original embeddings $h_k$ and $h_c$ will not be aligned, and therefore, neither will be $\bar{h}_c$ and $h_c$. Hence, their concatenation can be discriminatively used to identify when a joint has no good keypoint candidate for the target joint type. 

\section{Experiments}
\begin{table*}[ht!]
\begin{center}
\small
\setlength\tabcolsep{2pt}
\begin{tabular}{c|l|ccc|cc|c|c|c|c|c|c}
\toprule
\# &Method& Group.   & Type Agnostic  & Type Encoding  & Transformer & Pos. Encoding  & AP & $\text{AP}^{50}$ & $\text{AP}^{75}$  &  $\text{AP}^{M}$  &  $\text{AP}^{L}$ \\
\hline
 1 & Offsets + keypoint match.\cite{centernet}  &    	&      &  	    &   &   & 65.3 & 86.4  & 71.4  & 59.1  & 75.0\\ 
2 & AE \cite{hrnet, hourglass_ae} &  	&  &  &  &   &   67.1 &  86.2 & 73.0  & 61.5  & 76.1\\ 

\midrule
3 &Ours w/o K\&C transformer enc. & \checkmark  		&  	&      &  	    &   &     67.5  & 86.7 &     72.7    &   62.0    &  76.6\\ 
4 & Ours w/o K\&C transformer enc. & \checkmark  	&   \checkmark &       	    &   &   &       67.5& 86.8     & 72.9 & 60.8      & 77.3 \\ 
5 &Ours w/o K\&C transformer enc. & \checkmark  	&   \checkmark & \checkmark   	    &   &   &      67.9 & 87.4 &  73.2    &61.4 & 77.4\\ 
\midrule
6 & Ours & \checkmark  	&   \checkmark & \checkmark  &  \checkmark &     &     68.4 &    87.5 &      73.9&        62.0 &     77.6\\ 
7 &Ours & \checkmark  	&   \checkmark & \checkmark  &  \checkmark & \checkmark    &    \textbf{68.6}       &  \textbf{87.6} & \textbf{74.1} & \textbf{62.0} & \textbf{78.0}\\ 

%
\bottomrule
\end{tabular}
\caption{Ablation study on the \textbf{COCO2017 val} split.}\label{tab:main_ablation}
\end{center}
\vspace{-20pt}
\end{table*}
In this section, we detail the experimental evaluation of our method. We divide it into ablation studies and comparison to state-of-the-art on two large-scale public datasets. For implementation details, we refer the reader to the supplementary material.

\subsection{Datasets and Evaluation Metrics}
\noindent{\textbf{COCO Keypoint Detection.}} The COCO dataset \cite{coco_dataset} is a large-scale benchmark containing large variety of every-day life situations. It contains over 200,000 images and 17 keypoints annotations for more than 250,000 human instances, which are split in approximately 150,000, 80,000 and 20,000 instances are for training,  testing and, validation, respectively. We train our models on the \textit{train2017} split only, perform our ablation studies on \textit{val2017}, and report our final results on the \textit{test-dev2017} split.

\noindent{\textbf{CrowdPose.}} The CrowdPose dataset \cite{crowdpose} is a challenging benchmark with the goal to evaluate the robustness of methods in crowded scenes. Unlike COCO, in which the majority of images contain few instances, the \textit{crowd index} in CrowdPose follows a uniform distribution \cite{crowdpose}. The dataset contains a total of 20,000 images and a total of 80,000 instances annotated with 14 keypoints. Images are split in a ratio 5:4:1 for training, validation, and testing. Following \cite{hrnet}, we train our model on the train and validation splits combined, and report the final performance on the test set.

\noindent{\textbf{Evaluation metrics.}} The aforementioned datasets use average precision (AP) as their main metric. AP computation is based on the Object Keypoint Similarity (OKS)~\cite{coco_dataset} score among detected and ground truth poses. AP is the result of average precision scores for OKS thresholds $0.50, 0.55..., 0.90, 0.95$. We also report AP for thresholds $0.5$ and $0.75$, namely, $\text{AP}^{50}$ and $\text{AP}^{75}$. In addition, for COCO we report $\text{AP}^L$ and $\text{AP}^M$, which corresponds to AP over medium and large-sized instances respectively. For CrowdPose, we also report $\text{AP}^E$, $\text{AP}^M$, $\text{AP}^H$, which stands for AP scores over {easy}, {medium} and {hard} instances, according to dataset annotations.

\subsection{Ablation Study}
To determine the individual contribution of each of our model's main components, we perform an ablation study on the COCO val2017 split, 
with  HRNet32 backbone and input size 512x512. All results are reported with flip-testing, following \cite{hrnet,hourglass_ae, hgg}, and without top-down refinement. 

\noindent{\textbf{Baselines.}} CenterGroup can be naturally compared to two alternative frameworks. First, associative embeddings~\cite{hourglass_ae} (Tab. \ref{tab:main_ablation}, row \#1), since they are the method used originally by our keypoint detection network \cite{hrnet}.  Second, one-shot or offset-based methods \cite{centernet, spm}, which also use person center predictions, but use offset regression to obtain the final results. 
For a fair comparison, we reimplement \cite{centernet} with our HigherHRNet backbone and report its performance for its strongest variant, which predicts keypoint heatmaps, center heatmaps and center offsets, and matches centers to their closest predicted keypoint (Tab. \ref{tab:main_ablation}, row \#2).

\noindent{\textbf{Grouping Module.}} We consider our model without the transformer encoder to isolate the effect of our Grouping Module. 
%
%
We compare three versions of it.
In the first one, the attention head corresponding to the prediction of each keypoint type is only allowed to attend over keypoints of the same type that our keypoint detection network detects, and therefore is not able to overcome joint type mistakes made by the detector. 
This setting (Tab. \ref{tab:main_ablation}, row \#3) already outperforms our baselines, which confirms the superiority of CenterGroup over AE-based grouping and offset-based methods. 
In rows \#4 and \#5, we allow each head to attend over keypoints from the entire set of predicted heatmaps, and refer to them as \textit{type-agnostic}. In row \#5, we further use type encoding in the attention computation, as explained in Section \ref{section:routing}, and 
observe that they significantly improve upon type-agnostic grouping.


\noindent{\textbf{Feature encoding.}} In rows \#6 and \#7 of Table \ref{tab:main_ablation}, we further analyze the effect of using the keypoint and center transformer encoding before the routing module. 
This yields a significant performance boost, which confirms the importance of encoding long-range interactions between keypoints. Further enhancing the initial embeddings with positional encodings allows the transformer to explicitly use spatial information and gives up to 0.4 AP points of improvement for large persons.

\noindent{\textbf{Loss terms.}} We also assess the importance of our additional center and visibility classification losses, the results can be found in Table \ref{tab:abl_loss}. Without them, we score our predicted poses by directly assigning them the confidence of its predicted center from heatmaps. We observe that replacing the heatmap score with the classification score obtained from our transformer's embeddings (row \#2) already provides a significant boost. We then experiment with either using a simple MLP over those features to predict the visibility of every keypoint (row \#3), compared to using our attention-based model shown in Figure \ref{fig:routing}, results in row \#4. We observe that both yield a significant improvement, but our attention-based model performs best.

\noindent{\textbf{Runtime analysis.}} In Table \ref{tab:runtime}, we report the overall speed of our method when compared to our baselines. All models are run on the same machine with a single NVIDIA RTX5000 GPU, with batch size 1 and flip testing. We report: grouping runtime, i.e., all computations after keypoint detection, which in our case includes the transformer and grouping attention forward pass, and the overall runtime, which always adds 126ms corresponding to HigherHRNet. The overall runtime of CenterGroup is similar to \cite{centernet}, while we get significantly better results. Compared to AE-based grouping, our keypoint attention grouping is over 6x faster.


\begin{table}
\begin{center}
\resizebox{\columnwidth}{!}{
\begin{tabular}{c|c|c|c|c|c|c}
\toprule
\#&Class. Cent. & Vis. w/ MLP & Vis. w/ Attn. & AP & AP\textsuperscript{M} & AP\textsuperscript{L} \\
\midrule
1& &  & & 66.5 & 61.4 & 75.5 \\
2&\checkmark &  & & 67.1 & 61.0 & 76.2 \\
3&\checkmark & \checkmark & & 68.2 & 61.8 & 77.5 \\
4&\checkmark&  & \checkmark &  \textbf{68.6} & \textbf{62.0} & \textbf{78.0}\\
\bottomrule
\end{tabular}
}
\end{center}
\vspace{-12pt}
\caption{Ablation study on loss terms.} 
\label{tab:abl_loss}
\end{table}


\begin{table}
\begin{center}
\resizebox{\columnwidth}{!}{
\begin{tabular}{c|c|c|c|c|c}
\toprule
Method & Group. Time (ms) &  Time (ms) & AP & AP\textsuperscript{M} & AP\textsuperscript{L} \\
\midrule
Offsets + match\cite{centernet} & \textbf{20} & \textbf{146} & 65.3 & 59.1 & 75.0 \\
AE\cite{hourglass_ae} & 327 & 453 & 67.1 & 60.7 & 76.0 \\
\hline
Ours&   52 & 178 & \textbf{68.6} & \textbf{62.0} & \textbf{78.0}\\
\bottomrule
\end{tabular}
}
\end{center}
\vspace{-12pt}
\caption{Runtime analysis of different grouping methods.} 
\label{tab:runtime}
\end{table}

\subsection{Benchmark Evaluation}
\noindent{\textbf{COCO Keypoint Detection.}}
\begin{table}
\begin{center}
\resizebox{\columnwidth}{!}{
\begin{tabular}{l|c|c|c|c|c}
\toprule
Method & AP & AP\textsuperscript{50} & AP\textsuperscript{75} & AP\textsuperscript{M} & AP\textsuperscript{L} \\
\midrule
\multicolumn{6}{c}{Top-down methods} \\
\midrule
Mask-RCNN \cite{mask_rcnn}  & 63.1 & 87.3 & 68.7 & 57.8 & 71.4 \\
G-RMI \cite{g_rmi} &  64.0 & 85.5 & 71.3 & 62.3 & 70.0 \\
Integral Pose Regression \cite{integral} & 67.8 & 88.2 & 74.8 & 63.9 & 74.0\\
CPN \cite{cpn}& 72.1 & 91.4 & 80.0 & 68.7 & 77.2\\
RMPE \cite{rmpe}& 72.3 & 86.1 & 79.1 & 68.0 & 78.6 \\
CFN \cite{coarsefine}& 72.6 & 86.1 & 69.7 & \textbf{78.3} & 64.1\\
SimpleBaseline \cite{simple_baselines} & 73.7 & 91.9 & 81.1 & 70.3 & 80.0 \\
HRNet-W48 \cite{hrnet_original}& \textbf{75.5} & \textbf{92.5} & \textbf{83.3} & 71.9 & \textbf{81.5} \\
\midrule
\multicolumn{6}{c}{Bottom-up methods} \\
\midrule
OpenPose* \cite{openpose} & 61.8 & 84.9 & 67.5 & 57.1 & 68.2  \\
Hourglass*\textsuperscript{+} \cite{hourglass_ae} & 65.5 & 86.8 & 72.3 & 60.6 & 72.6  \\
PifPaf\cite{pifpaf} & 66.7 & - & - & 62.4 & 72.9 \\
SPM*\textsuperscript{+} \cite{spm} & 66.9 & 88.5 & 72.9 & 62.6 & 73.1 \\
HGG\textsuperscript{+} \cite{hgg}& 67.6 & 85.1 & 73.7 & 62.7 & 74.6 \\
PersonLab\textsuperscript{+} \cite{personlab} & 68.7 & 89.0 & 75.4 & 66.6 & 75.8 \\
HrHRNet-W32\cite{hrnet}  & 66.4 & 87.5 & 72.8 & 61.2 & 74.2 \\
HrHRNet-W48\cite{hrnet} & 68.4 & 88.2 & 75.1 & 64.4 & 74.2\\
HrHRNet-W48\textsuperscript{+} \cite{hrnet} & 70.5 & 89.3 & 77.2 & 66.6 & 75.8 \\
\hline
Ours w/ HrHRNet-W32 & 67.6 & 88.7 & 73.6 & 61.9 & 75.6\\
Ours w/ HrHRNet-W48 &  69.6 & 89.7 & 76.0 & 64.9 & 76.3\\
Ours w/ HrHRNet-W32\textsuperscript{+} & 70.3 & 90.0 & 76.9 & 65.4 & \textbf{77.5} \\



Ours w/ HrHRNet-W48\textsuperscript{+} & \textbf{71.4} & \textbf{90.4} & \textbf{78.1} & \textbf{67.2} &  \textbf{77.5} \\
\bottomrule
\end{tabular}
}
\end{center}
\caption{Comparisons with state of the art methods on the  \textbf{COCO2017 test-dev} split. * means top-down refinement, and \textsuperscript{+} means multi-scale testing.}
\label{tab:coco_test}
\vspace{-12pt}
\end{table}
In Table \ref{tab:coco_test}, we compare CenterGroup against state-of-the-art methods on the COCO dataset. Our method achieves the best performance among all bottom-up methods, for both single and multi-scale testing, and outperforms HigherHRNet, which uses AE-based Grouping~\cite{hourglass_ae}, by approximately 1 AP. We observe that our achievements are most significant in AP\textsuperscript{L}. This can be explained by the ability of our attention module to capture long-range interactions between joints that are far apart. Overall, strong results in COCO, combined with our faster inference speed, show that CenterGroup is a more efficient alternative to current bottom-up methods \cite{centernet}. We provide additional analysis in the supplementary material.

\noindent{\textbf{CrowdPose.}}
In Table \ref{tab:crowdpose_test}, we show the test-set results for our model trained on CrowdPose. Unlike COCO, where top-down methods show superior performance, bottom-up methods  outperform their top-down counterparts in CrowdPose, since this dataset is focused on much more challenging  images with severe occlusions. In this setting, CenterGroup shows its full potential and obtains state-of-the-art performance among all methods by 1.8 AP points.
Most importantly, our improvement is most significant in the hard regime (AP\textsuperscript{H}), where we improve upon state-of-the-art by 2.4 and 2.6  AP points for single and multi-scale testing, respectively.

This proves that our end-to-end learnable formulation does benefit from being trained on a dataset in which occlusions are common, and results in better generalization to new, challenging images. 
Overall, we show that our end-to-end trainable method can outperform top-down and bottom-up approaches on difficult scenarios with severe occlusions, where reasoning about keypoint detection and grouping jointly has a clear benefit.

\begin{table}
\begin{center}
\resizebox{\columnwidth}{!}{
\begin{tabular}{l|c|c|c|c|c|c}
\toprule
Method & AP & AP\textsuperscript{50} & AP\textsuperscript{75} & AP\textsuperscript{E} & AP\textsuperscript{M} & AP\textsuperscript{H} \\
\midrule
\multicolumn{7}{c}{Top-down methods} \\
\midrule
Mask-RCNN \cite{mask_rcnn} & 57.2 & 83.5 & 60.3 & 69.4 & 57.9 & 45.8 \\
SimpleBaseline \cite{simple_baselines}  & 60.8 & 81.4 & 65.7 & 71.4 & 61.2 & 51.2 \\
AlphaPose  \cite{rmpe} & 61.0 & 81.3 & 66.0 & 71.2 & 61.4 & 51.1 \\
\midrule
\multicolumn{7}{c}{Top-down with refinement} \\
\midrule
SPPE \cite{crowdpose}  & 66.0 & 84.2 & 71.5 & 75.5 & 66.3 & 57.4 \\
\midrule
\multicolumn{7}{c}{Bottom-up methods} \\
\midrule
OpenPose*\cite{openpose} & - & - & - & 62.7 & 48.7 & 32.3 \\
HrHRNet-W48\cite{hrnet} & 65.9 & 86.4 & 70.6& 73.3 & 66.5 & 57.9\\ 
HrHRNet-W48\textsuperscript{+} \cite{hrnet} & 67.6 & 87.4 & 72.6& 75.8 & 68.1 & 58.9\\ 
\midrule
Ours w/ HrHRNet-W48 & 67.6 & 87.7 & 72.7 & 73.9 & 68.2 & 60.3\\ 
Ours w/ HrHRNet-W48\textsuperscript{+}  & \textbf{70.0} & \textbf{88.9} & \textbf{75.1} & \textbf{76.8} & \textbf{70.7} & \textbf{62.2}\\ 

\bottomrule
\end{tabular}
}
\end{center}
\caption{Comparisions with state of the art methods on the \textbf{CrowdPose test} set. Superscripts E, M, H mean easy, medium and hard.\textsuperscript{+} means multi-scale test, and * means top-down refinement. }
\label{tab:crowdpose_test}
\vspace{-15pt}
\end{table}

\section{Conclusion}

We have proposed an end-to-end attention-based framework for bottom-up human pose estimation. We have demonstrated that CenterGroup has better performance than existing state-of-the-art methods, particularly in crowded images, while being significantly more efficient. 
We hope that our approach will inspire future work to explore the potential of attention mechanisms, as well as general learning-based alternatives to optimization-based grouping for bottom-up human pose estimation.

\noindent{\bf {Acknowledgments.}}
\small{
This project was partially funded by the Sofja Kovalevskaja Award of the Humboldt Foundation and by the German Federal Ministry of Education and Research (BMBF) under Grant No. 01IS18036B. The authors of this work take full responsibility for its content.
}


{\small
\bibliographystyle{ieee_fullname}
\bibliography{ms}
}
\clearpage
\appendix
\normalsize
\noindent{\LARGE{\textbf{Supplementary Material}}}

\maketitle

\thispagestyle{empty}
\global\csname @topnum\endcsname 0
\global\csname @botnum\endcsname 0





\section{Extended COCO Comparison}
\begin{table*}[ht!]
\begin{center}
\small
\setlength\tabcolsep{2pt}
\begin{tabular}{l|c|c|c|c|ccccc}
\toprule
Method& Backbone & Grouping & Input size & \# Params & AP & $\text{AP}^{50}$ & $\text{AP}^{75}$  &  $\text{AP}^{M}$  &  $\text{AP}^{L}$ \\
\midrule
\multicolumn{10}{c}{w/o multi-scale test} \\
\midrule
OpenPose* \cite{openpose} & -- & Greedy decoding w/ optimization &  --  & -- & 61.8 & 84.9 & 67.5 & 57.1 & 68.2  \\
AE* \cite{hourglass_ae} & Hourglass& Greedy decoding w/ optimization & 512 & 277.8M & 62.8 & 84.6 & 69.2 & 57.5 & 70.4  \\
PersonLab\cite{personlab} & ResNet152 &Greedy decoding& 1401 & 68.7M  &   66.5 & 88.0 & 72.6 & 62.4 & 72.3 \\
PifPaf\cite{pifpaf} & -- &Greedy decoding w/ optimization   & --  & -- &  66.7 & - & - & 62.4 & 72.9 \\
HigherHRNet\cite{hrnet, hourglass_ae} & HRNet-W32 &Greedy decoding w/ optimization & 512 & 28.6M  & 66.4 & 87.5 & 72.8 & 61.2 & 74.2 \\
HigherHRNet\cite{hrnet, hourglass_ae}  & HRNet-W48 &Greedy decoding w/ optimization  & 640 & 63.8M  & 68.4 & 88.2 & 75.1 & 64.4 & 74.2\\
\hline
Ours  & HRNet-W32 & Attention & 512 & 30.3M  &  67.6 & 88.7 & 73.6 & 61.9 & 75.6\\
Ours  & HRNet-W48 & Attention & 640 & 65.5M  & \textbf{69.6} & \textbf{89.7} & \textbf{76.0} & \textbf{64.9} & \textbf{76.3}\\
\midrule
\multicolumn{10}{c}{w/ multi-scale test} \\
\midrule
AE* \cite{hourglass_ae} & Hourglass & Greedy decoding w/ optimization & 512 & 277.8M & 65.5 & 86.8 & 72.3 & 60.6 & 72.6  \\
SPM* \cite{spm} &Hourglass & Offsets (One-shot) & 512 & 277.8M   &  66.9 & 88.5 & 72.9 & 62.6 & 73.1 \\
HGG \cite{hgg}&Hourglass &Graph Network + Graclus clustering \cite{graclus} & 512 & --& 67.6 & 85.1 & 73.7 & 62.7 & 74.6 \\
PersonLab \cite{personlab} &ResNet152 & Greedy decoding & 1401 & 68.7M  & 68.7 & 89.0 & 75.4 & 66.6 & 75.8 \\
HrHRNet-W48 \cite{hrnet}  &HRNet-W48 & Greedy decoding w/ optimization & 640 & 63.8M   & 70.5 & 89.3 & 77.2 & 66.6 & 75.8 \\
\hline
Ours &HRNet-W32& Attention & 512 & 30.3M  & 70.3 & 90.0 & 76.9 & 65.4 & \textbf{77.5} \\
Ours &HRNet-W48& Attention & 640 & 65.5M & \textbf{71.4} & \textbf{90.4} & \textbf{78.1} & \textbf{67.2} &  \textbf{77.5} \\

\bottomrule
\end{tabular}
\caption{Comparison of published bottom-up methods on the \textbf{COCO2017 test-dev} split. * means top-down refinement. \textit{w/ optimization} refers to the use of bipartite matching solvers during inference.}\label{tab:coco_bu}
\end{center}
\end{table*}
In Table \ref{tab:coco_bu}, we provide a detailed comparison of CenterGroup against published bottom-up approaches on the COCO test-dev dataset. For each method, we specify its backbone network, grouping procedure, input size, and parameter count. We observe that most top-performing methods rely on greedy decoding schemes, which often involve optimization in the form of solving a sequence of bipartite matching problems. Alternatively, SPM \cite{spm} uses offsets, but relies on top-down refinement to achieve competitive results \footnote{i.e. it applies a single person pose estimation model over the predicted poses.}, and HGG\cite{hgg} uses a hierarchical clustering algorithm that operates on the output of graph network predictions. 

CenterGroup outperforms all previous methods with our proposed attention-based grouping module, which does not rely on optimization and is end-to-end trainable. Note that this module only introduces a slight increase in the number of parameters with respect to HigherHRNet\cite{hrnet}, and combined with our keypoint detector, yields a model with significantly fewer parameters than other methods. 

Regarding performance, we note that the increase in accuracy is most significant for large persons, where our improvement is of 2.1 AP points for single-scale, and 1.7 for multi-scale, which can be explained by the ability of our transformer to capture relationships among distant joints in the image. Overall, it outperforms the current state-of-the-art method, HigherHRNet\cite{hrnet}  by approximately 1.2 AP for single-scale and 0.9 AP for multi-scale, while having the exact same backbone and input size, and being 2.5x faster, which confirms CenterGroup's increased efficiency.
\section{Matching Centers}
In order to train our grouping module, we need to determine which detected centers in the image correspond to a ground truth pose. As explained in Section 5.4 in the main paper, this allows us to define a target $y_{\text{center}}^c$ for every detected center $c\in \mathcal{C}$ indicating whether it represents a ground truth pose (i.e., $y_{\text{center}}^c=1$) or not ($y_{\text{center}}^c=0$). These labels are used to train our center classification module. 
Moreover, for those detected centers that do correspond to a ground truth pose, we obtain the visibility of their corresponding keypoints as well as the locations of those that are visible by simply using the annotations of the ground truth center that the detected center is matched with.

In order to determine correspondences between detected centers ($\mathcal{C}$) and ground truth centers ($\mathcal{P}$), we compute the euclidean distance between every $c\in \mathcal{C}$ and $\bar{c}\in \mathcal{P}$, and normalize it by the scale of $\bar{c}$, $s_{\bar{c}}$:
\begin{equation} \label{match_distance}
    \text{dist}(c, \bar{c}):=exp\left (- \frac{ || \text{loc}_c - \text{loc}_{\bar{c}}||^2}{2s_{\bar{c}}*k^2} \right )
\end{equation}
where $k$ is a fixed constant set to 0.15\footnote{This number is determined by increasing by 50\% the constant that the COCO dataset uses for hip joints for OKS computation.}, and the scale $s_{\bar{c}}$ is computed as $0.53$ multiplied by $\bar{c}$'s bounding box height and width, following \cite{crowdpose}. This formula is adapted from the OKS metric, and simply normalizes distances between 0 and 1 by using a pre-defined standard deviation that depends on the object size.

With the distances from Equation \ref{match_distance}, we define an instance of a bipartite matching problem. For every $c\in \mathcal{C}$ and $\bar{c}\in \mathcal{P}$, their corresponding cost $\text{cost}(c, \bar{c}) := 1 - \text{dist}(c, \bar{c})$, whenever  $\text{dist}(c, \bar{c})<0.5$ and $\infty$ otherwise. 
We obtain matches between centers and ground truth centers by solving the problem with the hungarian algorithm, similarly to \cite{detr}. Note that running this algorithm takes on average significantly less than $1$ms since the cost matrix is, at most, of size 20x30, and therefore it adds no significant computational burden. Additionally, note that this procedure is only necessary at training time in order define ground truth assignments. At test-time, as explained in the main paper, we do not require any form of optimization.



\section{Implementation Details}
\subsection{Training}
We pretrain our backbone and keypoint detection module following HigherHRNet \cite{hrnet}. We then randomly initialize our encoding and grouping modules and train our entire model end-to-end for $27,000$ iterations with batch size 130, which corresponds to approximately $~$50 epochs on COCO, and $~$270 epochs on CrowdPose, and use learning rate linear warm-up during the first $1,000$ iterations\cite{imagenet1hour}. We use an Adam optimizer \cite{adam} with learning rate set to $1e-5$ for pretrained layers and $3e-4$ for the remaining parts of the network, which we drop by a factor of 10 at  10,000 and 20,000 iterations. In addition, use use automatic mixed precision for training \cite{mixed_precision}, which reduces the memory requirements by approximately half, and allows training on 4 NVIDIA RTX6000 with 24GB of RAM memory in approximately 24 hours. We observe that our training loss shows high stability and allows training with mixed precision without any divergence problems, in contrast to Associative Embeddings\cite{hourglass_ae}.
For data augmentation, we use the same techniques as \cite{hrnet}, which include random flipping, rotation, scale variation, and generating a random crop of size 512x512, when using an HRNet32 backbone,  or 640x640 when using an HRNet48 backbone.

We add one grouping module at the output of every transformer encoder block and compute the location, visibility and center losses, and then average them over the output of every transformer encoder block. Loss terms are balanced as follows: the heatmap loss, $\mathcal{L}_{\text{heatmap}}$ is weighted by factor 10, the location loss, $\mathcal{L}_{\text{loc}}$ is averaged over all visible keypoints in the image and weighted by 0.02, the center and visibility losses, $\mathcal{L}_{\text{vis}}$ and $\mathcal{L}_{\text{center}}$, are both weighted by factor 1. The overall set of weights is determined by ensuring that each loss term has a comparable magnitude. 
\subsection{Inference}
At inference, we resize images to preserve their aspect ratio and have their shorter side of size 512 if using a HRNet32 backbone, or 640 if using HRNet48. Following \cite{hrnet}, predicted heatmaps are upsampled to full image resolution. We then extract peaks by applying heatmap Non-Maximum Suppression (NMS) with a max-pooling kernel of size 5x5 for keypoints and 17x17 for person centers, and select all peaks that either have score over 0.01 or are within the top-5 scoring peaks in the heatmap.

For every predicted center $c\in \mathcal{C}$, we build its pose by assigning it the keypoints with highest attention score according to the attention score corresponding to every type, as explained in Section 4.2 in the main paper. Formally, given center $c\in \mathcal{C}$ the location of each of its joint types $i\in \{1, \dots, J\}$ is determined as:
\begin{equation}
    \widehat{\text{loc}}_c^i = \argmax_{k\in \mathcal{K}} \text{attn}_i(c, k)
\end{equation}
In order to score the resulting poses, we use the predicted visibility scores for every keypoint, $\widehat{\text{vis}}_c^i$, as well as the predicted probability that center $c$ represents a true positive center, $\hat{y}_\text{center}^c$, as follows:
\begin{equation}
\text{score}_c = \begin{cases}
\text{avg} \left (\{ \widehat{\text{vis}}_c^i \mid \widehat{\text{vis}}_c^i \geq 0.5 \}_{i=1}^J \right ) &\text{if } \hat{y}_\text{center}^c \geq 0.5\\
\hat{y}_\text{center}^c &\text{otherwise}
\end{cases}
\end{equation}
Intuitively, since visibility scores are only computed for those centers such that $y_{\text{center}}^c=1$ during training (i.e. matched centers), we only use them whenever our network predicts centers to represent true pose centers with probability over 0.5. In that case, the overall pose score is the average visibility confidence score of keypoints that are predicted to be visible (i.e.,  $\widehat{\text{vis}}_c^i \geq 0.5$).

Unlike \cite{hourglass_ae, spm, openpose}, we do not perform top-down refinement, nor ensembling \cite{multiposenet}, and all results are reported with flip-testing as it is common practice \cite{hrnet_original, hourglass_ae, personlab}. For postprocessing, following \cite{hourglass_ae, hrnet}, keypoint coordinates are shifted by 0.25 towards the contiguous second maximal activation in each heatmap, to account for quantization errors.

\subsection{Exact Architecture}
\begin{figure}[t]
\begin{center}
   \includegraphics[width=0.5\linewidth]{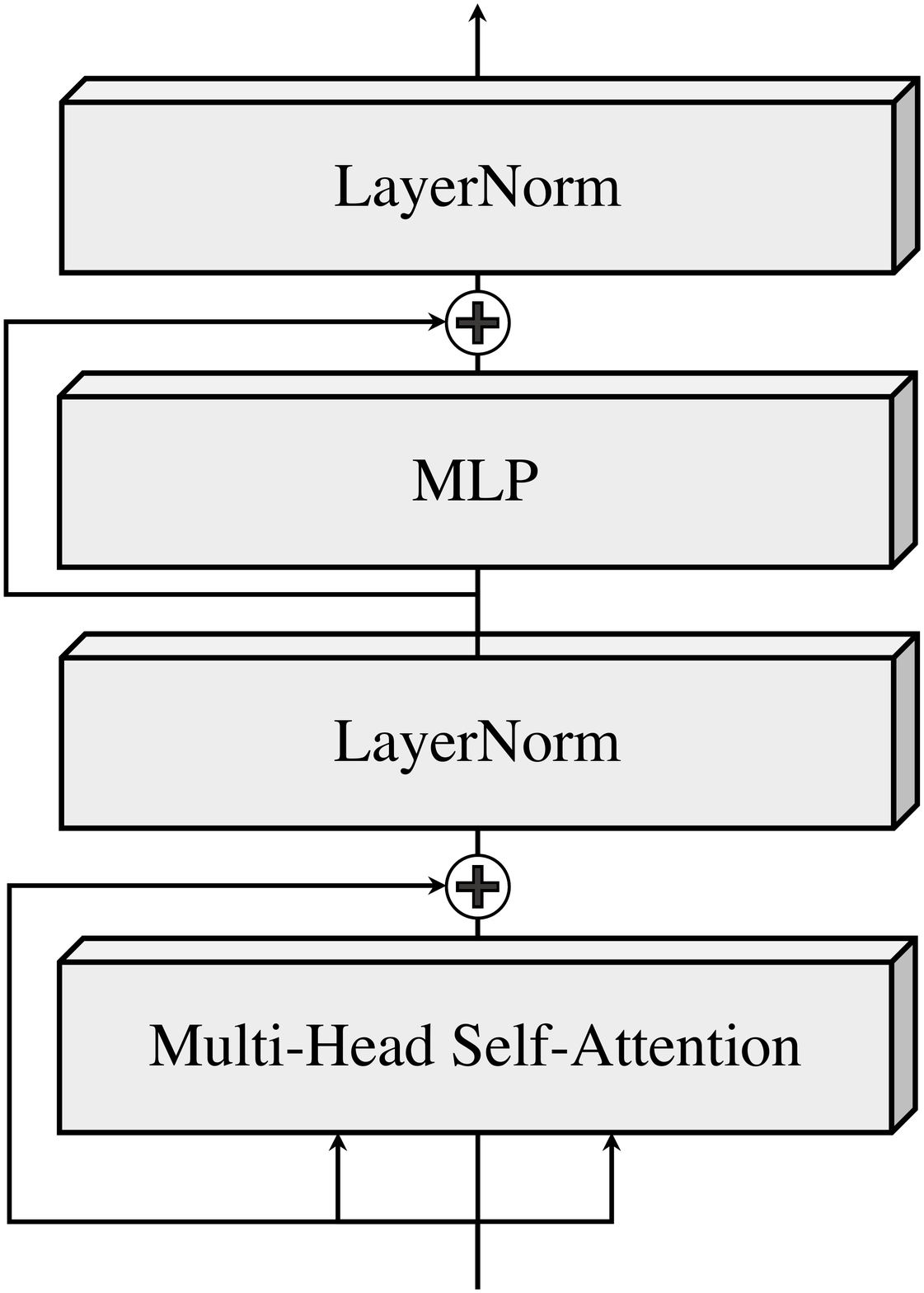}
\end{center}
   \caption{Overview of the architecture of a Transformer Encoder. }
\label{fig:transformer_encoder}
\vspace{-10pt}
\end{figure}

Our keypoint detection network is minimally modified from HigherHRNet, as explained in Section 5.2 in the main paper. Our newly added modules include an additional residual block and a multi-layer perceptron (MLP) to generate initial keypoint and person features, a transformer encoder and the grouping module. Our transformer encoder has 3 blocks, each with input dimension 128, 4 self-attention heads and MLP hidden dimension set to 512. We found no significant performance benefits from further increasing the transformer's size. The architecture of each transformer encoder block is not modified from the original one \cite{transformer}, and shown in Figure \ref{fig:transformer_encoder}.

\begin{table}
\begin{center}
\resizebox{0.7\columnwidth}{!}{
\begin{tabular}{l|c}
\toprule
Layer Name   & \# Parameters \\
\midrule
\multicolumn{2}{c}{Keypoint Detection} \\
\midrule
Backbone & 28.5M  (63.7M)\\
Keypoint Heads & 110K \\
\midrule
\multicolumn{2}{c}{Encoding} \\
\midrule
Residual Block & 595K \\
Initial MLP & 33K \\
Transformer Encoder & 594K \\
\midrule
\multicolumn{2}{c}{Grouping} \\
\midrule
Multi-Head Attention & 420K \\
$\text{MLP}_{\text{center}}$ & 33K \\
$\text{MLP}_{\text{vis}}$  & 41K \\
\midrule
\multicolumn{2}{c}{Overall} \\
\midrule
-- & 30.3M \\
\bottomrule
\end{tabular}
}
\end{center}
\vspace{-12pt}
\caption{Parameter count breakdown among components in each stage of our model's pipeline. For the backbone, 28.5M refers to a HRNet32 backbone, and 63.7M refers to a HRNet48. Note that the overall number of parameters of our proposed encoding and grouping modules combined is relatively small, at 1.7M.} 
\label{tab:params}
\end{table}

All of the MLPs in the grouping module, as well as the one generating the transformer's input contain two hidden layers. We detail the number of parameters of each component in Table \ref{tab:params}. The overall parameter count of our proposed keypoint encoding and grouping module is below 2M, which is relatively small, and only accounts for $<$6$\%$ (resp. $<$3$\%$) of the overall count when using an HRNet32 (resp. HRNet48) backbone. 

\section{Qualitative Results}
\subsection{Qualitative Examples}
\begin{figure*}[ht]
 \begin{center}
 \vspace{-0.5cm}
      \begin{subfigure}[b]{0.23\textwidth}
          \centering
          \includegraphics[width=\textwidth]{LaTeX/figures/qual_images/000000000872.jpg}
          \includegraphics[width=\textwidth]{LaTeX/figures/qual_images/000000229601.jpg}
          \includegraphics[width=\textwidth]{LaTeX/figures/qual_images/000000363666.jpg}            
          \includegraphics[width=\textwidth]{LaTeX/figures/qual_images/000000567640.jpg}                
          \includegraphics[width=\textwidth]{LaTeX/figures/qual_images/000000147740.jpg}
          \includegraphics[width=\textwidth]{LaTeX/figures/qual_images/000000005193.jpg}
         \includegraphics[width=\textwidth]{LaTeX/figures/qual_images/000000473219.jpg}     
           \caption{Input Image}
      \end{subfigure}
      \begin{subfigure}[b]{0.23\textwidth}
          \centering
          \includegraphics[width=\textwidth]{LaTeX/figures/qual_results/hrnet_872.jpg}
          \includegraphics[width=\textwidth]{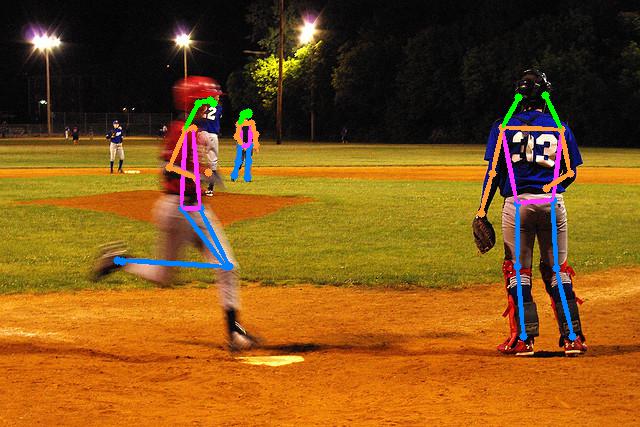}
          \includegraphics[width=\textwidth]{LaTeX/figures/qual_results/hrnet_363666.jpg}            
          \includegraphics[width=\textwidth]{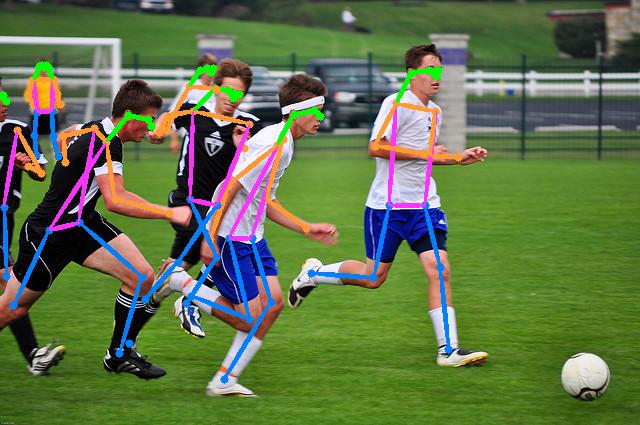}            
          \includegraphics[width=\textwidth]{LaTeX/figures/qual_results/hrnet_147740.jpg}
          \includegraphics[width=\textwidth]{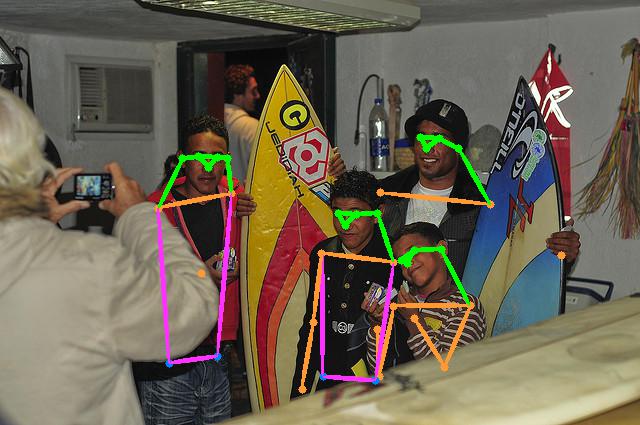}
         \includegraphics[width=\textwidth]{LaTeX/figures/qual_results/hrnet_473219.jpg}          
           \caption{HigherHRNet \cite{hrnet}}
      \end{subfigure}
      \begin{subfigure}[b]{0.23\textwidth}
          \centering
          \includegraphics[width=\textwidth]{LaTeX/figures/qual_results/centernet_872.jpg}
          \includegraphics[width=\textwidth]{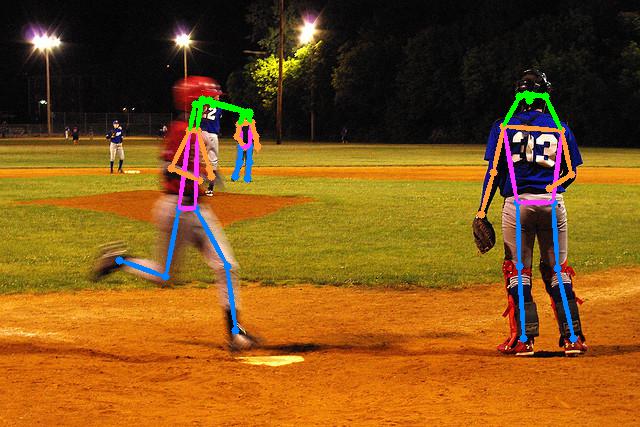}
          \includegraphics[width=\textwidth]{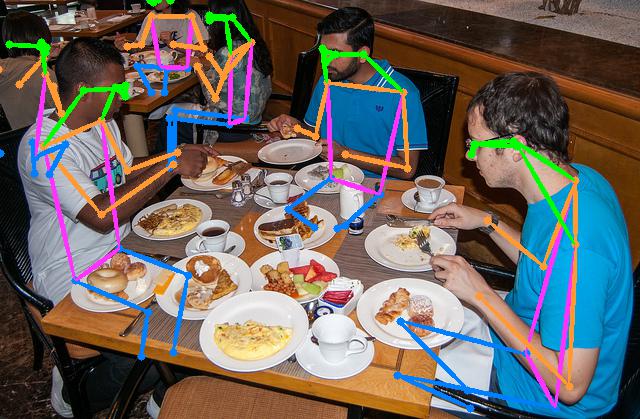}            
          \includegraphics[width=\textwidth]{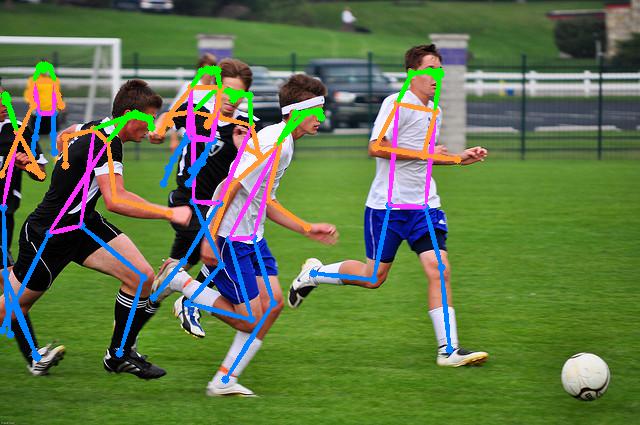}            
          \includegraphics[width=\textwidth]{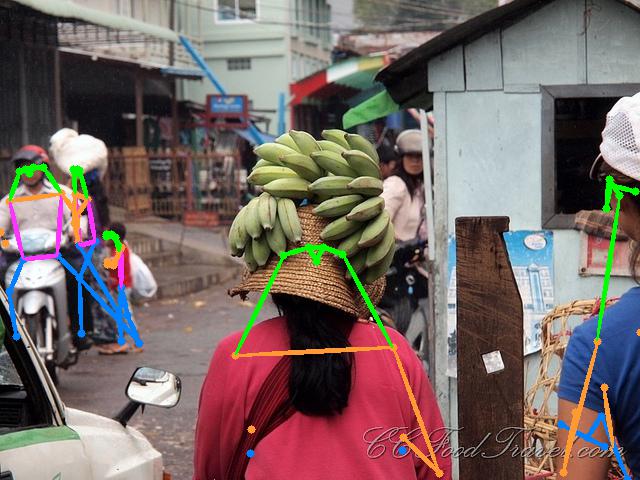}
          \includegraphics[width=\textwidth]{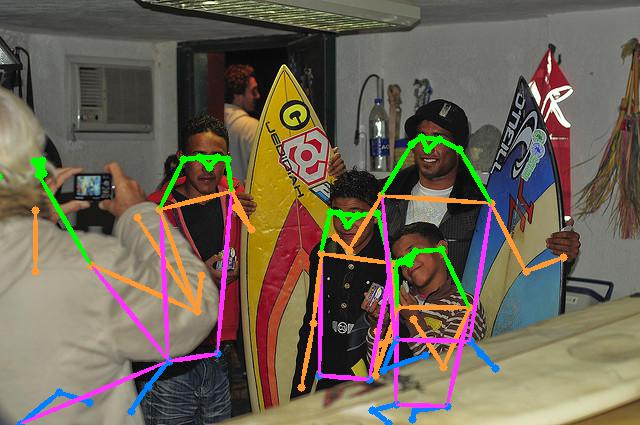}
         \includegraphics[width=\textwidth]{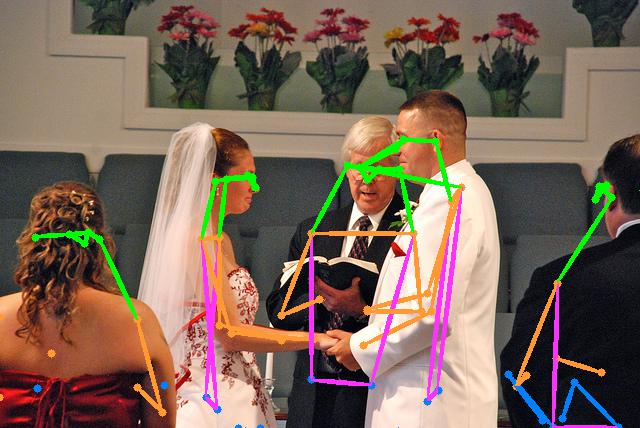}         
           \caption{CenterNet\cite{centernet}}
      \end{subfigure}
      \begin{subfigure}[b]{0.23\textwidth}
          \centering
          \includegraphics[width=\textwidth]{LaTeX/figures/qual_results/ours_872.jpg}
          \includegraphics[width=\textwidth]{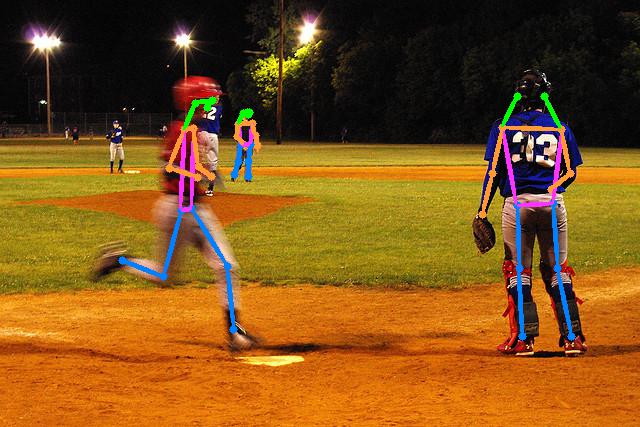}
          \includegraphics[width=\textwidth]{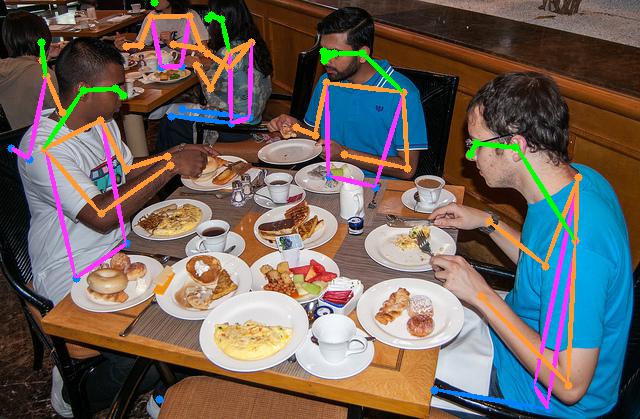}
          \includegraphics[width=\textwidth]{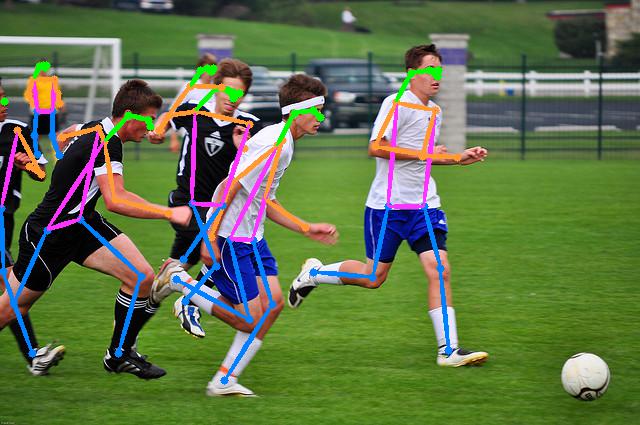}
          \includegraphics[width=\textwidth]{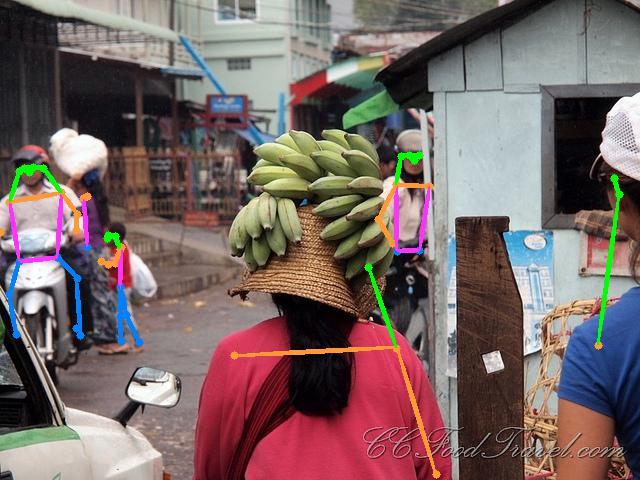}
          \includegraphics[width=\textwidth]{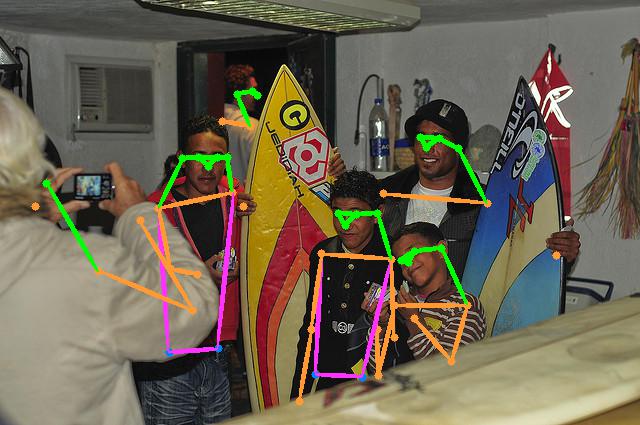}         
         \includegraphics[width=\textwidth]{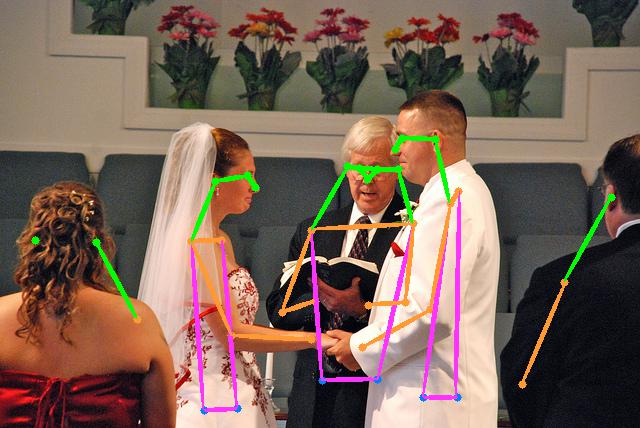}         
           \caption{Ours}
           \label{fig:of}
      \end{subfigure}
 \caption{Qualitative examples of our method's performance in comparison to HigherHRNet\cite{hrnet} and CenterNet\cite{centernet}. Best viewed in color and in a screen.}
 \label{fig:qual_results}
 \vspace{0.1cm}
 \end{center}
 \end{figure*}

In Figure \ref{fig:qual_results}, we visualize results produced by our method in comparison to those from our baselines: HigherHRNet\cite{hrnet} and CenterNet \cite{centernet}. As explained in the main paper, we reimplement CenterNet to use an HRNet\cite{hrnet_original} backbone and HigherHRNet's scale-aware heatmaps \cite{hrnet} for keypoint heatmap regression for a fair comparison. 

We observe that our method's performance is robust under severe occlusion and challenging conditions. In comparison, CenterNet often fails whenever there is significant overlap among different poses, as can be seen in rows 1, 4, 5, 6 and 7. Moreover, since it always predicts joint locations for a given pose regardless of whether they are visible or not, it often hallucinates joints and produces unfeasible pose estimates (all rows). 

HigherHRNet generally does a better job at grouping, as can be seen in rows 1, 4, 5, and 6, but this comes at a significantly increased computational cost of 2.5x inference time. Moreover, we observe that it tends to miss or assign very low confidence to large-sized poses (rows 2, 4, 5, 6).

Our method, instead, has a runtime inference time comparable to CenterNet's, due to its fast optimization-free test-time procedure, and has increased robustness where our baselines fail. Namely, it performs well in images with heavy occlusion, and, due to its ability to capture long-rage connections with our attention mechanism, it does not struggle with large-sized poses.

 \begin{figure*}[ht]
 \begin{center}
 \vspace{-0.5cm}
      \begin{subfigure}[b]{0.3\textwidth}
          \centering
          \includegraphics[width=\textwidth]{LaTeX/figures/qual_images/000000000872.jpg}
          \includegraphics[width=\textwidth]{LaTeX/figures/qual_images/000000229601.jpg}
          \includegraphics[width=\textwidth]{LaTeX/figures/qual_images/000000363666.jpg}            
          \includegraphics[width=\textwidth]{LaTeX/figures/qual_images/000000567640.jpg}                
          \includegraphics[width=\textwidth]{LaTeX/figures/qual_images/000000147740.jpg}
           \caption{Input Image}
      \end{subfigure}
      \begin{subfigure}[b]{0.3\textwidth}
          \centering
          \includegraphics[width=\textwidth]{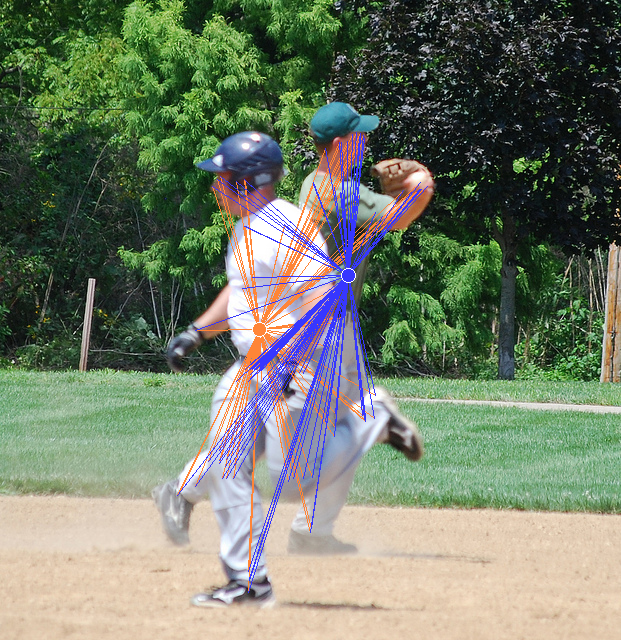}
          \includegraphics[width=\textwidth]{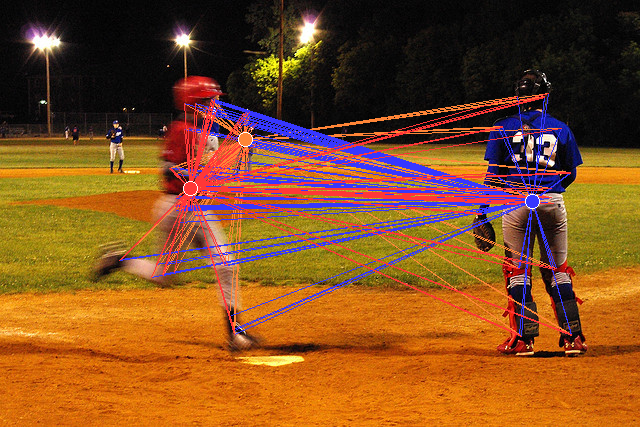}
          \includegraphics[width=\textwidth]{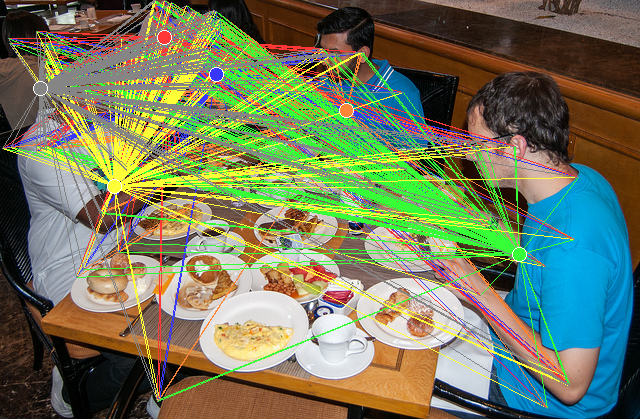}
          \includegraphics[width=\textwidth]{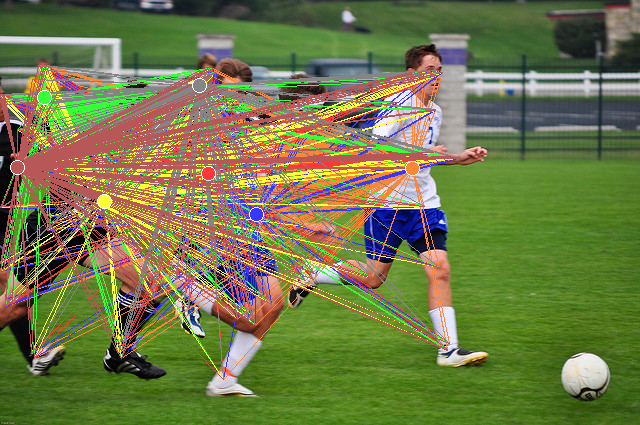}
          \includegraphics[width=\textwidth]{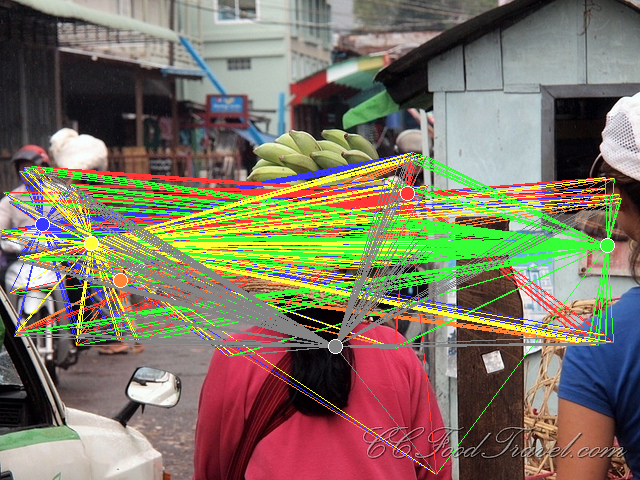}
           \caption{Center Keypoint Connections }
      \end{subfigure}
      \begin{subfigure}[b]{0.3\textwidth}
          \centering
          \includegraphics[width=\textwidth]{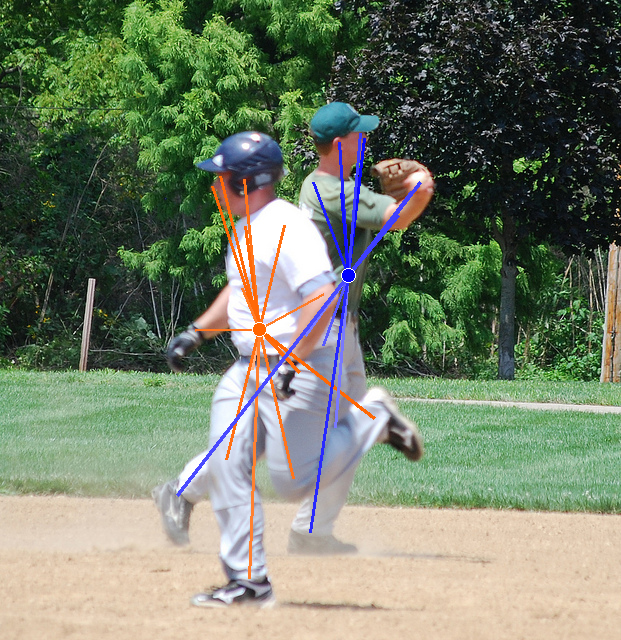}
          \includegraphics[width=\textwidth]{LaTeX/figures/attn_viz_results/attn_conns_im_ix_229601.png}
          \includegraphics[width=\textwidth]{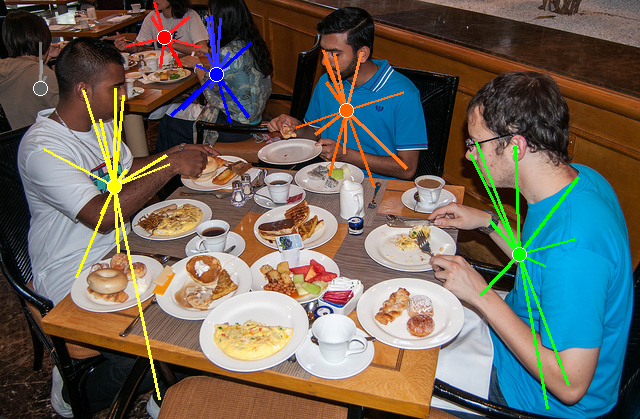}
          \includegraphics[width=\textwidth]{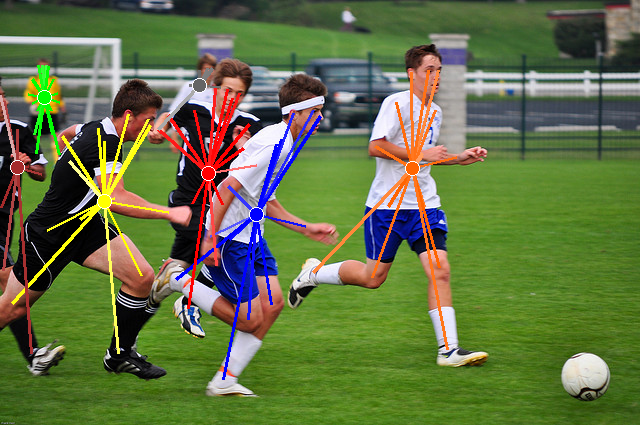}
          \includegraphics[width=\textwidth]{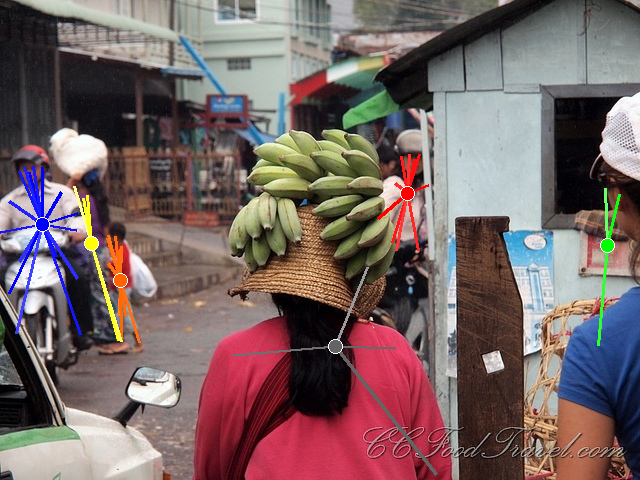}
           \caption{Predicted Attention Scores}
      \end{subfigure}
 \vspace{-7pt}
 \caption{Visualization of predicted attention scores by our grouping module. In (b) we show all pairwise connections between detected keypoints and centers classified as true positives. In (c) we show all final attention scores predicted with attention weight over 0.5 and as visible. The attention weight is color-coded in the color's intensity. Best viewed in color and in a screen.}
 \label{fig:attn1}
 \vspace{0.1cm}
 \end{center}
 \end{figure*}

  \begin{figure*}[ht]
 \begin{center}
 \vspace{-0.5cm}
      \begin{subfigure}[b]{0.3\textwidth}
          \centering
          \includegraphics[width=\textwidth]{LaTeX/figures/qual_images/000000005193.jpg}
          \includegraphics[width=\textwidth]{LaTeX/figures/qual_images/000000473219.jpg}     
           \caption{Input Image}
      \end{subfigure}
      \begin{subfigure}[b]{0.3\textwidth}
          \centering
          \includegraphics[width=\textwidth]{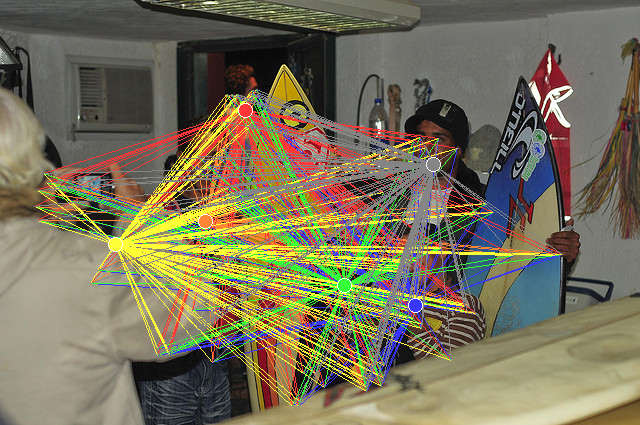}
          \includegraphics[width=\textwidth]{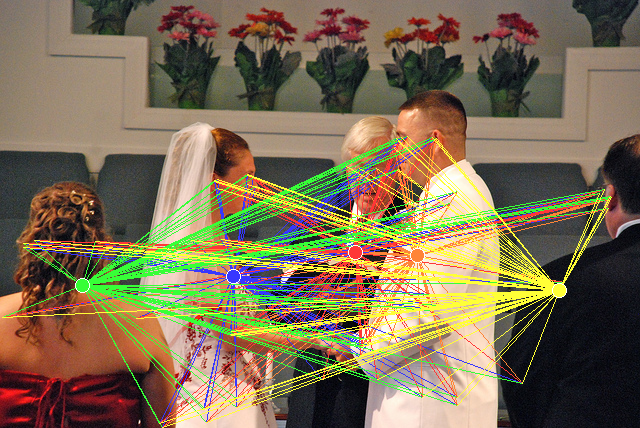}
           \caption{Center Keypoint Connections }
      \end{subfigure}
      \begin{subfigure}[b]{0.3\textwidth}
          \centering
          \includegraphics[width=\textwidth]{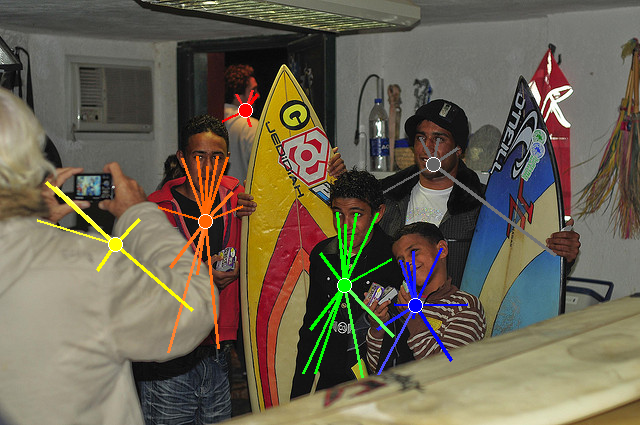}
          \includegraphics[width=\textwidth]{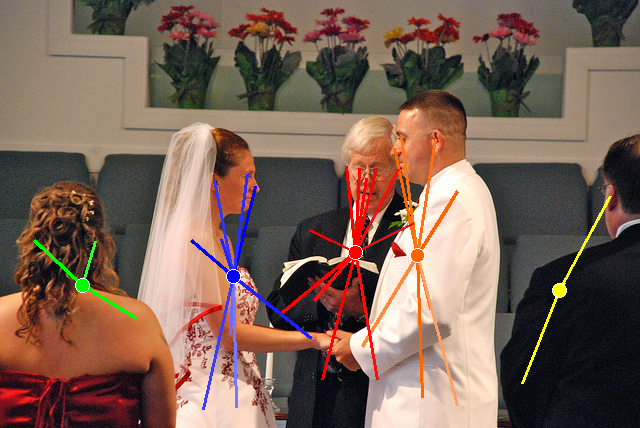}
           \caption{Predicted Attention Scores}
      \end{subfigure}
 \vspace{-7pt}
 \caption{Visualization of predicted attention scores by our grouping module. In (b) we show all pairwise connections between detected keypoints and centers classified as true positives. In (c) we show all final attention scores predicted with attention weight over 0.5 and as visible. The attention weight is color-coded in the color's intensity. Best viewed in color and in a screen.}
 \label{fig:attn2}
 \vspace{0.1cm}
 \end{center}
 \end{figure*}

\subsection{Visualizing Attention Activations}
In Figures \ref{fig:attn1} and \ref{fig:attn2} we visualize the attention output scores with which the results in Figure \ref{fig:qual_results} were obtained. We observe that despite the large amount of keypoints over which each center attends, particularly in crowded scenes, attention scores are heavily concentrated over a small subset of keypoints, for each center. Indeed, most attention scores for a given type have magnitude over 0.95\%, which can be seen from the dark color of most lines. This can be explained due to our loss formulation: to achieve low training error, our model must concentrate attention weights in the most promising keypoint locations, as otherwise it'd incur in large L1 loss values.  Overall, Figures \ref{fig:attn1} and \ref{fig:attn2} show how our model is able to consider a large number of center-keypoint association candidates but still focus on those keypoints belonging to each pose, even in highly challenging scenarios.






\end{document}